\documentclass[10pt,twocolumn,letterpaper]{article}

\usepackage[pagenumbers]{wacv} 

\usepackage[dvipsnames,table]{xcolor}
\usepackage{graphicx}
\usepackage{makecell}
\usepackage{pifont}
\usepackage{multirow}
\usepackage{amsmath}
\usepackage{arydshln}
\usepackage{booktabs}
\usepackage{nicematrix}
\usepackage{float}
\usepackage[normalem]{ulem}
\usepackage{chngcntr}
\newcommand{\tablestyle}[2]{\setlength{\tabcolsep}{#1}\renewcommand{\arraystretch}{#2}\centering\footnotesize}

\newif\ifclean

\newcommand{\cc}[1]{\ifclean \else {\color{Green}{#1}}\fi}

\newcommand{\PM}[1]{\ifclean{#1}\else {\color{orange}{#1}}\fi}
\newcommand{\edit}[1]{\ifclean{#1}\else {\color{Green}{#1}}\fi}

\newcommand{\stat}[1]{\ifclean{#1}\else {\color{violet}{#1}}\fi}

\newcommand{\dataset}{WhereToChange} 
\newcommand{\datanum}{1162\ } 
\newcommand{\htc}{HowToChange} 

\newcommand{\method}{$\textsc{SPOC}$}

\newcommand{\task}{spatially-progressing\ }
\newcommand{\greencheck}{{\color{PineGreen}\ding{51}}} 
\newcommand{\redcross}{{\color{red}\ding{55}}}  

\newcommand{\custompar}[1]{
  \par
  \vspace{0.5pt}
  \noindent\textbf{#1}
}

\newcommand\blfootnote[1]{%
  \begingroup
  \renewcommand\thefootnote{}\footnote{#1}%
  \addtocounter{footnote}{-1}%
  \endgroup
}

\definecolor{LighterGray}{gray}{0.93}
\definecolor{DarkerGray}{gray}{0.73}
\newcolumntype{?}{!{\vrule width 1pt}}
\newcolumntype{g}{>{\columncolor{DarkerGray}}c}

\newif\iftransf  
\transftrue  

\newif\ifsam

\newif\iflossconstraint

\cleantrue  

\definecolor{wacvblue}{rgb}{0.21,0.49,0.74}
\usepackage[pagebackref,breaklinks,colorlinks,allcolors=wacvblue]{hyperref}
\usepackage[accsupp]{axessibility}  

\def\wacvPaperID{604} 
\def\confName{WACV}
\def\confYear{2026}

\title{SPOC: Spatially-Progressing Object State Change Segmentation in Video}

\author{Priyanka Mandikal \quad Tushar Nagarajan \quad Alex Stoken \quad Zihui Xue \quad Kristen Grauman\\
The University of Texas at Austin\\
{\tt\small mandikal@utexas.edu}
}

\begin{document}
\maketitle
\begin{abstract}
Object state changes in video reveal critical 
cues
about human and agent activity.  However, existing methods are limited to temporal localization of when the object is in its initial state 
(e.g., cheese block) 
versus when it has completed a state change 
(e.g., grated cheese), 
offering no insight into where the change is unfolding.
We propose to deepen the problem by introducing the 
spatially-progressing object state change segmentation
task. The goal is to segment at the pixel-level those regions of an object that are actionable and those that are transformed. 
We show that state-of-the-art VLMs and video segmentation methods struggle at this task, underscoring its difficulty and novelty.
As an initial baseline, we
design a VLM-based pseudo-labeling approach, state-change dynamics constraints, and a novel WhereToChange benchmark built on in-the-wild Internet videos.  Experiments on two datasets validate both the challenge of the new task as well as the promise of our model for localizing exactly where and how fast objects are changing in video.
We further demonstrate useful implications for tracking activity progress to benefit robotic agents.
Overall, our work positions spatial OSC segmentation as a new frontier task for video understanding: one that challenges current SOTA methods and invites the community to build more robust, state-change-sensitive representations.
\blfootnote{Project webpage: \url{https://vision.cs.utexas.edu/projects/spoc-spatially-progressing-osc}}
\end{abstract}

\vspace{-1em}
\section{Introduction}
\label{sec:intro}

Understanding object state changes (OSC) is crucial for various applications in computer vision and robotics. OSC refers to the dynamic process where objects transition from one state to another, such as a whole apple being sliced, a potato being mashed, 
or a piece of cloth being ironed.
Accurately capturing and analyzing these transitions is essential for tasks that require real-time decision-making and action planning. For instance, in human activity videos, 
precise OSC understanding 
could enable AR/MR assistants
to monitor the progress of a user's activities and prompt timely instructions and interventions. 
In robotics, 
it
would enable robots to perform complex tasks, such as cooking or assembly, by determining 
when and where to 
act next.

\begin{figure}[t]
    \centering
    \includegraphics[width=1.0\linewidth]{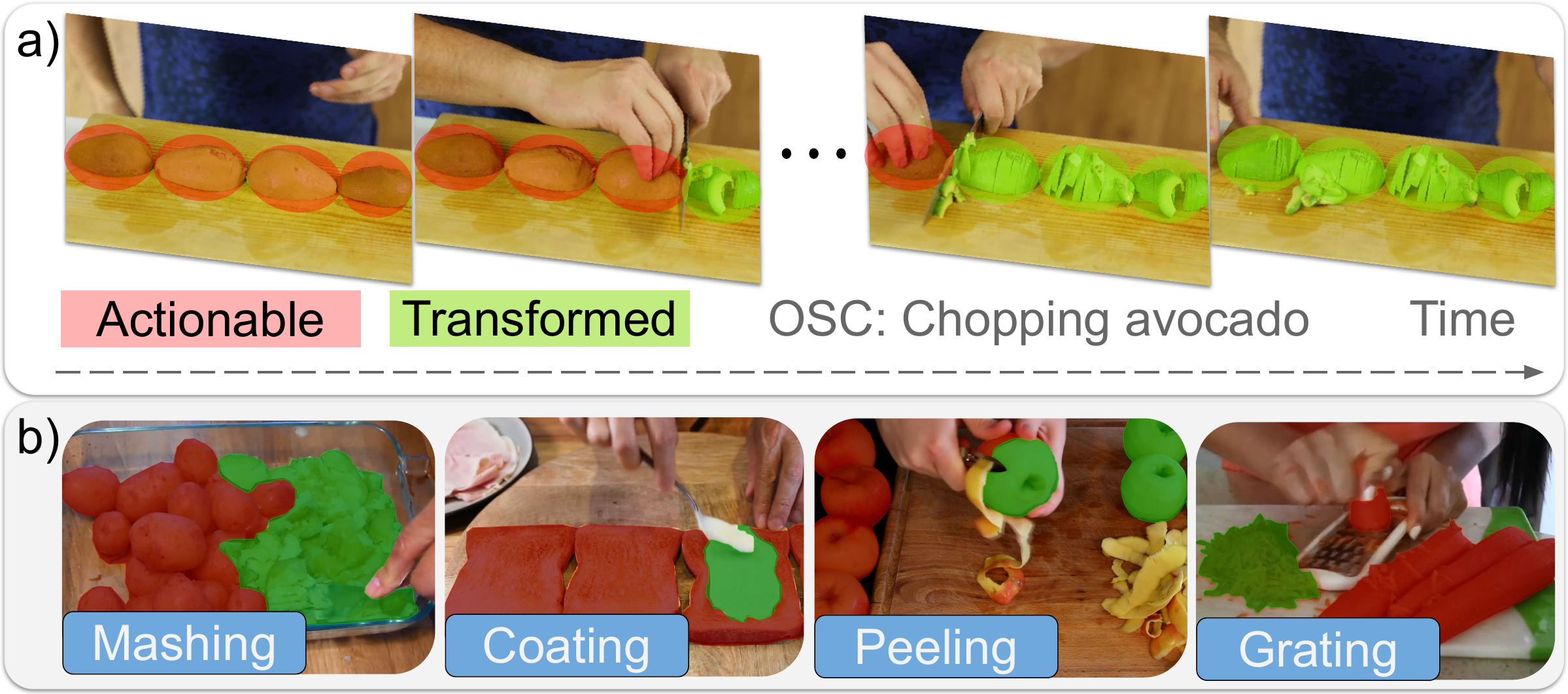}
    \caption{
    a) An illustration of the \task video OSC segmentation problem:
    With time, the regions within an object undergo a progressive state-change from actionable to transformed.
    b) A diverse set of spatially-progressing segmentations for different state-change activities.
     Red is actionable; green is transformed.
     }
    \label{fig:intro}
    \vspace{-2em}
\end{figure}

Current approaches for OSC typically classify video frames as initial, transitioning, or end states~\cite{souvcek2022look,souvcek2022multi,xue2023vidosc} or segment entire objects undergoing OSCs~\cite{Yu_2023_vscos,tokmakov2023vost}. However,  existing work makes an important assumption that all parts of the interacted objects change simultaneously and uniformly over time.
We observe that in reality, an OSC often also \emph{spatially progresses during an action}: (1) state changes occur to specific localized object regions over time, and (2) parts of objects can undergo changes non-uniformly. 
For example, in Fig.~\ref{fig:intro}, 
the cheese is grated over dough,
and parts of the potato mixture get mashed over time. 
Though not possible with existing methods, these pixel-level spatial distinctions are essential for planning and execution of any such state-changing process---such as a robot deciding exactly \emph{where} to act next, or an AI assistant capturing a smart-glasses user's progress on their task.

To address this fundamental limitation, we introduce a novel task formulation:
\task OSC segmentation (\method). 
Given an OSC, we formally categorize the object regions into one of two states---\textit{actionable} or \textit{transformed}.  For example, in Fig.~\ref{fig:intro}a,
the grated pieces of the cheese are transformed, while the yet-to-be-grated chunk is actionable.
Given a video, the goal is to actively segment and label these regions in each frame as the object undergoes the state change transition. 
These changes are typically irreversible, meaning the object cannot return to its original state—an inherent characteristic of common OSCs~\cite{souvcek2022look,xue2023vidosc}.
Note that not all OSCs are spatially-progressing.  We define a spatially-progressing OSC as a
state change that 
sequentially affects different regions of the object.
This includes a wide variety of human activities like grating, peeling, shredding, painting and so on, and excludes those where the whole object undergoes state change uniformly, like grilling, frying, or blending\footnote{We consider cooking-related OSCs following prior work~\cite{souvcek2022look,xue2023vidosc} due to their real-world value, complexity, wide diversity, and data availability; however our approach is generalizable to other domains.}.

Our approach copes with the intrinsic challenges of spatial OSC, where objects can undergo dramatic transformations in color, texture, shape, topology, and size, often bearing little resemblance to their original form. These challenges thwart direct application of today's video object segmentation models~\cite{zhou2022maskclip,ren2024gsam}, as we will see in results, and call for new training resources and objectives to be developed.

Equipped with this new task formulation, we propose two innovations to tackle spatially-progressing OSC. First, we design a pipeline to acquire large-scale training data by pseudo-labeling human activity videos with vision–language models (VLMs). By combining segmentation models with language–image embeddings, we generate weak labels sufficient to bootstrap training. Second, we introduce state-change dynamics constraints—causal ordering and ambiguity resolution—that enforce temporal consistency and correct ambiguous VLM predictions. Leveraging both, we train a video model that labels mask proposals with actionable or transformed states across frames.

Complementing our task and model contributions, we present \dataset, the first benchmark for \task video OSC, with fine-grained intra-object state-change segmentations (Table~\ref{tab:dataset}). Derived from  HowToChange~\citep{xue2023vidosc} and VOST~\cite{tokmakov2023vost}, our dataset includes human-annotated OSC segmentations for \datanum human-activity video clips, spanning 
210 state changes and 102 unique objects.
Experiments on \dataset\ 
demonstrate that our model outperforms adapted state-of-the-art methods on this new benchmark,
and its spatial OSC signals provide useful progress curves for robotic reward shaping.
Yet the benchmark remains far from solved: even state-of-the-art VLMs (e.g., GPT-4o) struggle to classify state-changing parts consistently. This positions spatial OSC segmentation as a novel and challenging benchmark that pushes beyond current VLM and VOS capabilities, offering a testbed for advancing state-change–aware video object representations.

\section{Related Work}
\label{sec:rel_work}

\custompar{Object state changes}
Current approaches to OSC understanding fall into three main areas: classification~\citep{souvcek2022look,souvcek2022multi,xue2023vidosc}, temporal segmentation~\citep{Yu_2023_vscos,tokmakov2023vost}, and generation~\citep{soucek2024genhowto}.
Early OSC methods focused on image-level classification~\citep{misra2017red,nagarajan2018attributes,mit_states,purushwalkam2019task,mancini2021open,naeem2021learning,pham2021learning,li2022siamese}, later expanding to video-based approaches that leverage the dynamics of OSCs~\citep{alayrac2017joint,liu2017jointly,souvcek2022look}. The HowToChange dataset~\citep{xue2023vidosc} further extended classification to unseen object categories.
Recently, VOST~\citep{tokmakov2023vost} and VSCOS~\citep{yu2023video} introduced the task of segmenting objects undergoing state changes in videos,  
but only address the segmentation 
of the \textit{entire} object throughout the video. In contrast, we provide
spatial annotations of state changes both \textit{within} and \textit{across} object instances,
for a more detailed benchmark for video object segmentation. 
Additionally, our OSC dynamics operate at the individual object level rather than at a broader frame level.

\custompar{Video object segmentation}
Video object segmentation (VOS) aims to separate foreground objects from background at the pixel level. Most work focuses on semi-supervised VOS~\citep{yang2021aot,cheng2022xmem,yang2022deaot}, where target objects are tracked across a video given ground-truth masks in the first frame. Benchmarks such as DAVIS~\citep{Perazzi2016davis,Caelles_arXiv_2019_davis} and YouTube-VOS~\citep{xu2018youtubevos} emphasize objects that maintain structural integrity (e.g., people, cars, animals). More recent datasets include state-changing objects but treat them at the whole-object level~\citep{tokmakov2023vost,Yu_2023_vscos}. In contrast, our formulation and benchmark require intra-object spatio-temporal segmentation of state-change progress.

\custompar{Open-vocabulary segmentation}
Segment-Anything (SAM)~\citep{kirillov2023sam,ravi2024sam2} is a promptable segmentation model with zero-shot generalization to unfamiliar objects and images. While SAM originally worked with click or bounding-box prompts, several subsequent works integrate SAM with textual prompts in an open-vocabulary setting~\citep{ren2024gsam,cheng2023samtrack,cheng2023deva,shen2024rosa}, to segment all object instances belonging to the categories described.
These methods fall short for our task since they are often unable to distinguish between different states of the object, as we show in results.
However, our strategy to generate large-scale pseudo-labels from VLMs~\cite{clip} adhering to specific state-change constraints can infuse this object state-sensitive information into model training, yielding significantly stronger performance.

\custompar{Vision and language}
CLIP~\cite{clip} has been widely used for downstream tasks such as generation and open-vocabulary recognition~\cite{Lin_2023_CVPR,Patashnik_2021_ICCV,xu2022odise,yu2023fcclip}. However, since it was trained on image–text pairs, CLIP lacks pixel-level detail, performing poorly when applied directly to bounding box or segmentation tasks~\cite{liang2023maskadaptclip,yu2023fcclip,zhong2022regionclip,zhou2022maskclip}. Extensions such as ROSA~\cite{shen2024rosa} adapt CLIP for local region–phrase grounding in narrational videos, while VidOSC~\cite{xue2023vidosc} uses CLIP to generate pseudo-labels for the HowToChange dataset, but only at the frame level.  
Our approach differs in two key ways: (1) we target \textit{intra-object} segmentation of state-change transitions, going beyond whole-object segmentation (ROSA~\cite{shen2024rosa}) or global temporal labels (VidOSC~\cite{xue2023vidosc}), and (2) we introduce novel object-level constraints that enforce state-change dynamics during pseudo-label generation.

\vspace*{-0.1in}
\section{Approach}
\label{sec:approach}

\begin{figure*}[t!]
    \centering
    \includegraphics[width=0.9\linewidth]{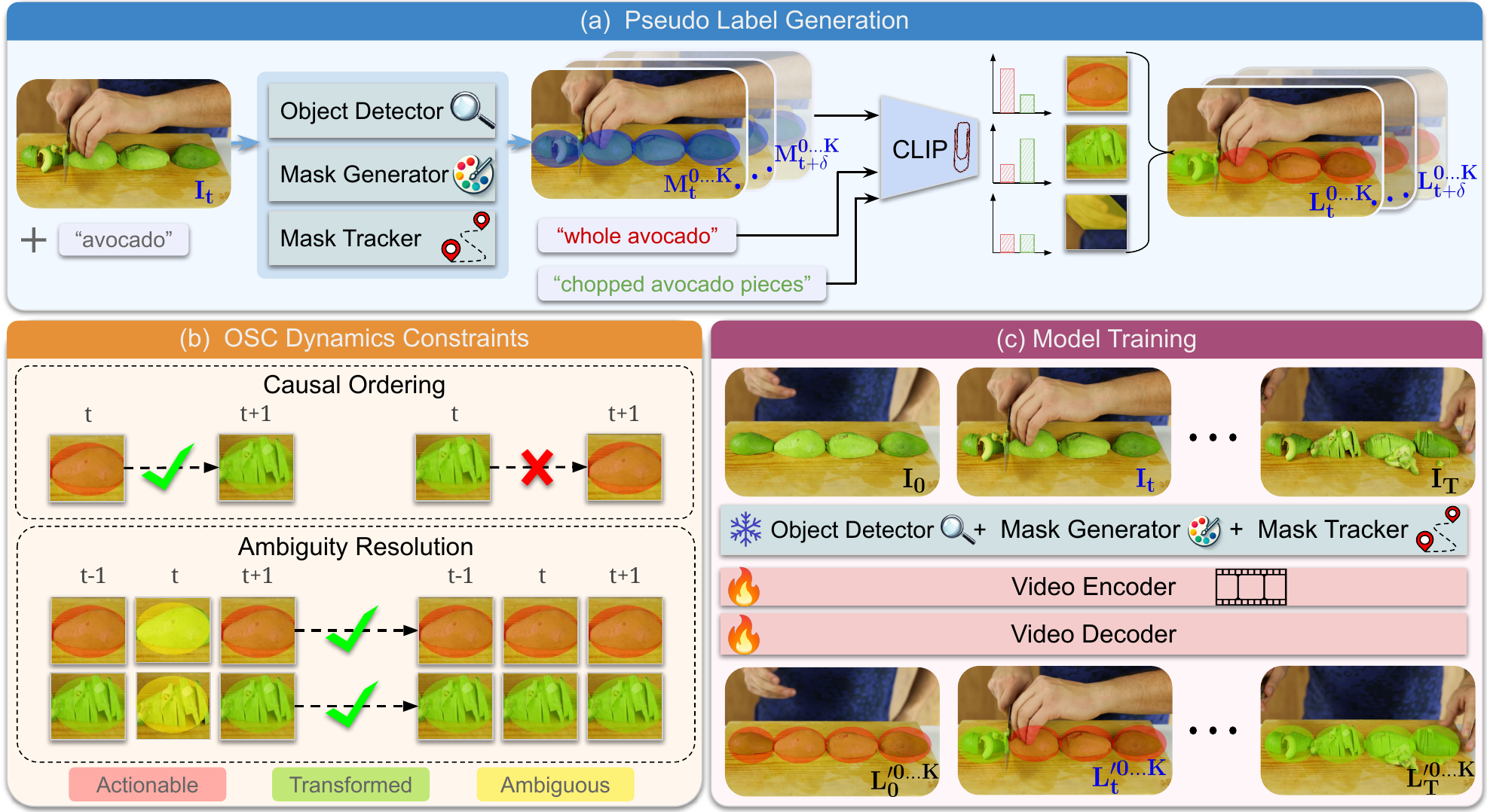}
    \caption{
    \textbf{Overview of \method{}.} 
    \textbf{a)} Pseudo-label generation (Sec.~\ref{subsec:pseudo_label_gen}): Given a video of a human performing a state-changing activity, we use off-the-shelf object detection~\cite{liu2023gdino}, mask generation~\cite{kirillov2023sam} and tracking models~\cite{yang2022deaot} to extract a set of region mask proposals $M^{0...K}_{t}$ for each frame $I_t$. We then use CLIP~\cite{clip} to apply similarity-score matching of visual region embeddings with textual state-description embeddings to obtain max-similarity pseudo-labels $L^{k}_t$ for each region. 
    \textbf{b)} OSC dynamics constraints (Sec.~\ref{subsec:constraints}): We refine the pseudo-labels by incorporating several important dynamics constraints that emphasize the temporal progression of state-change transitions while respecting their causal dynamics. 
    \iftransf
        \textbf{c)} Model training (Sec.~\ref{subsec:model_training}): Using our large-scale pseudo-labeled dataset, we train a video model to classify mask proposals into one of three classes: \textit{actionable}, \textit{transformed}, or \textit{background}. Evaluation is done  on manually labeled samples.
    \fi
    }
    \label{fig:overview}
    \vspace{-1em}
\end{figure*}

We present \method, a framework for spatially-progressing OSC segmentation. We present the problem formulation  (Sec.~\ref{subsec:problem_formulation}), followed by our approach to generate pseudo-labels to train our model (Sec.~\ref{subsec:pseudo_label_gen}) and
incorporate OSC dynamics constraints (Sec.~\ref{subsec:constraints}). Finally, we present our model-design and training details for state-change sensitive object segmentation (Sec.~\ref{subsec:model_training}).

\subsection{The Spatially-Progressing OSC Task}
\label{subsec:problem_formulation}

We cast the problem as a state-change dependent video segmentation task.
More formally, 
given a sequence of video frames \(\{I_1, I_2, \ldots, I_T\}\), where \(I_t\) represents the video frame at time-step \(t\), the objective is to generate segmentation masks \(\{L_t^0, L_t^1, \ldots, L_t^K\}\) for each frame \(I_t\) in the video sequence. Each mask \(L_t^k\) should accurately represent the spatial regions of the \(k\)-th object region in one of two states (\textit{actionable} or \textit{transformed})\footnote{Our inference formulation allows  predicting ``background" labels.} at time-step \(t\). 
The states are defined as follows: 
the actionable state (\(s_{\text{act}}\)) is the state of the object region before any change has taken place 
(e.g., cheese block),
and the transformed state (\(s_{\text{trf}}\)) is the state of the object region after transformation 
(e.g., grated cheese). 
All regions that do not yet adhere to the final state are actionable, providing
a clear binary formulation for our task.
Note that one object can have multiple regions simultaneously in different states.
Following prior OSC work~\cite{souvcek2022look,souvcek2022multi,xue2023vidosc}, we focus specifically on cooking activity videos, which include a
wide variety of complex and common irreversible OSCs,
supported by available datasets~\cite{miech2019howto100m, Damen2018EPICKITCHENS, grauman2022ego4d} (see Table~\ref{tab:htc_vost_big} for the full set).

\subsection{Pseudo-Label Generation}
\label{subsec:pseudo_label_gen}

Existing state-of-the-art open-set object detection methods~\cite{liu2023gdino,Cheng2024YOLOWorld} work well for broad object categories. However, identifying objects with specific fine-grained state attributes---especially those that drastically alter the object’s appearance---remains challenging.  To address this, we train a model specifically for state-change disambiguation, first leveraging VLMs in a novel approach to generate large-scale pseudo-labeled data,
 as follows (Fig.~\ref{fig:overview}a):

\custompar{Mask Proposal Generation}
Given a video frame \( I_t \) and OSC (e.g. chop avocado), we first detect and localize the relevant object by name (e.g., “avocado”) using an open-vocabulary object detector~\cite{ren2024gsam} to produce bounding boxes \( B_t^{0..K} \).
We assume that each video clip is associated with a single OSC---typical in in-the-wild videos~\cite{xue2023vidosc,grauman2022ego4d,Damen2018EPICKITCHENS}. 
Using these bounding boxes, we generate segmentation masks \( M_t^{0..K} \) for each detected object, leveraging a pre-trained segmentation model~\cite{ren2024gsam} to produce high-quality masks. These masks are then tracked~\cite{yang2022deaot} across frames to maintain consistency, capturing object state transitions over time. This tracking also enables the application of OSC dynamics constraints, as defined below.

\custompar{Mask Labeling} 
To distinguish between states $s_{\text{act}}$ and $s_{\text{trf}}$, we integrate the CLIP~\cite{clip} model. CLIP's vision-language embeddings are utilized to generate semantic labels through a thresholding process that aligns visual features of object regions with descriptive text annotations of the state change. To label a mask, we first obtain a vision embedding $z_v$ for the object region within the mask. Then, we compute textual embeddings for the two state descriptions 
(e.g., $\mathbb{T}_{act}$=``whole avocado", $\mathbb{T}_{trf}$=``chopped avocado pieces") to produce $z_{\mathbb{T}} \in \mathbb{R}^{d\times 2}$, where $d$ is the embedding dimension. 
These descriptions are automatically generated by few-shot prompting an LLM with the OSC verb and noun.
We then compute similarity scores between the vision-language embeddings by taking a simple dot product $\mathbf{S}=z_v \cdot z_{\mathbb{T}}$, where $\mathbf{S} \in \mathbb{R}^{2}$ is a similarity score matrix.
We can then obtain the pseudo-label for the object region as:

{\footnotesize
\begin{equation}
\hat y_t = 
\begin{cases} 
\text{Background} & \text{if } \mathbf{S}_{act}^t + \mathbf{S}_{trf}^t < \tau \\
\text{Ambiguous} & \text{elif } |\mathbf{S}_{act}^t - \mathbf{S}_{trf}^t| < \delta \\
\text{Actionable} &  \text{elif } \mathbf{S}_{act}^t > \mathbf{S}_{trf}^t \\
\text{Transformed} & \text{elif } \mathbf{S}_{trf}^t > \mathbf{S}_{act}^t.
\end{cases}
\label{eq:clip_thr}
\end{equation}
}

Here, $\delta$ is the threshold that separates object states, and 
$\tau$ differentiates between object states and background
(which helps filter out irrelevant object detections).
Similar to \cite{xue2023vidosc}, if the VLM 
strongly favors one state and the overall confidence is high, that state is assigned as the pseudo-label; otherwise, the case is marked ambiguous.
This process (illustrated in Fig.~\ref{fig:overview}a, right) generates preliminary pseudo-labels which we refine further as described next.

\subsection{OSC Dynamics Constraints}
\label{subsec:constraints}

While CLIP-based similarity matching provides an initial set of pseudo-labels to bootstrap training, these labels contain considerable noise. For example, labels for a specific object may erroneously transition from transformed to actionable state (impossible with irreversible OSCs), or regions initially marked as background may later be relabeled as the object of interest.
To effectively train our model for \task OSC segmentation, we first refine these pseudo-labels using a set of state-change dynamics constraints that capture the core aspects of state transitions.

We first define the variables in use. Let \( c \in \{ s_{\text{act}}, s_{\text{trf}}, s_{\text{amb}} \} \) denote the object-state class. Given a video with \( T \) frames, for each mask proposal \( k \) at time-step \( t \), let \( l_t^k \) represent the pseudo-label. We define \( \mathbb{S}_{\text{c}}^{k} = \{ t \in [1, T] : l_t^k = c \} \) as the set of time indices where proposal \( k \) is labeled as \( c \). With masklets corresponding to individual object instances, this formulation allows us to apply state-change dynamics constraints on pseudo-labels at the object-level, extending beyond frame-level application~\cite{souvcek2022look}.

\custompar{Causal ordering} 
We define a causality constraint to enforce a coherent progressive order among the labels at the \textit{object-level}. For a given tracked masklet $k$, the sequence of states should adhere to a progression where any instances of transformed states $s_{\text{trf}}$ appear after actionable states $s_{\text{act}}$.
To enforce this, we first compute their time-series mid-points:
   \[
   \text{mid}_{\text{act}} = \frac{\sum_{t \in \mathbb{S}_{\text{act}}^{k}} t}{|\mathbb{S}_{\text{act}}^{k}|}, \quad
   \text{mid}_{\text{trf}} = \frac{\sum_{t \in \mathbb{S}_{\text{trf}}^{k}} t}{|\mathbb{S}_{\text{trf}}^{k}|} 
   \]
If \( \mathbb{S}_{act}[-1] > \mathbb{S}_{trf}[0] \) (i.e., actionable indices follow transformed), 
we compute \( dist(\mathbb{S}_{act}[-1], mid_{act}) \) and \( dist(\mathbb{S}_{trf}[0], mid_{trf}) \). If the former is greater (indicating a potential mislabeled outlier in \( \mathbb{S}_{act} \)), we flip \( \mathbb{S}_{act}[-1] \); else flip \( \mathbb{S}_{trf}[0] \). 
We iteratively repeat until \( \mathbb{S}_{act}[-1] < \mathbb{S}_{trf}[0] \).
After this procedure, all 
indices in \( \mathbb{S}_{act} \) precede \( \mathbb{S}_{trf} \), enforcing a cohesive causal ordering in the pseudolabels.

\custompar{Ambiguity Resolution}
After enforcing causal ordering, we are still left with ambiguous pseudo-labels. 
For any ambiguous state $s_{\text{amb}}$ in the sequence, we assign it to either transformed or actionable class based on proximity:
   \[
   l_t^k\in{s_{\text{amb}}} \rightarrow
   \begin{cases} 
      s_{\text{act}}, & \text{if } |t - \max(S_{\text{act}}^{k})| < |t - \min(S_{\text{trf}}^{k})| \\
      s_{\text{trf}}, & \text{otherwise} 
   \end{cases}
   \]
where $t$ is the frame index with an ambiguous label. 
This re-labeling maintains continuity in state transitions, even for labels beyond the actionable-transformed boundaries.

By refining the label sequences with these dynamics constraints, we significantly reduce label noise, creating a more stable and logically consistent training set. Results show this refinement---which goes beyond traditional temporal smoothing---plays an important role in enhancing the reliability of pseudo-labels, enabling our model to learn from a structured representation of object state dynamics.

\subsection{Model Training}
\label{subsec:model_training}

\begin{figure}[t!]
    \centering
    \includegraphics[width=0.75\linewidth]{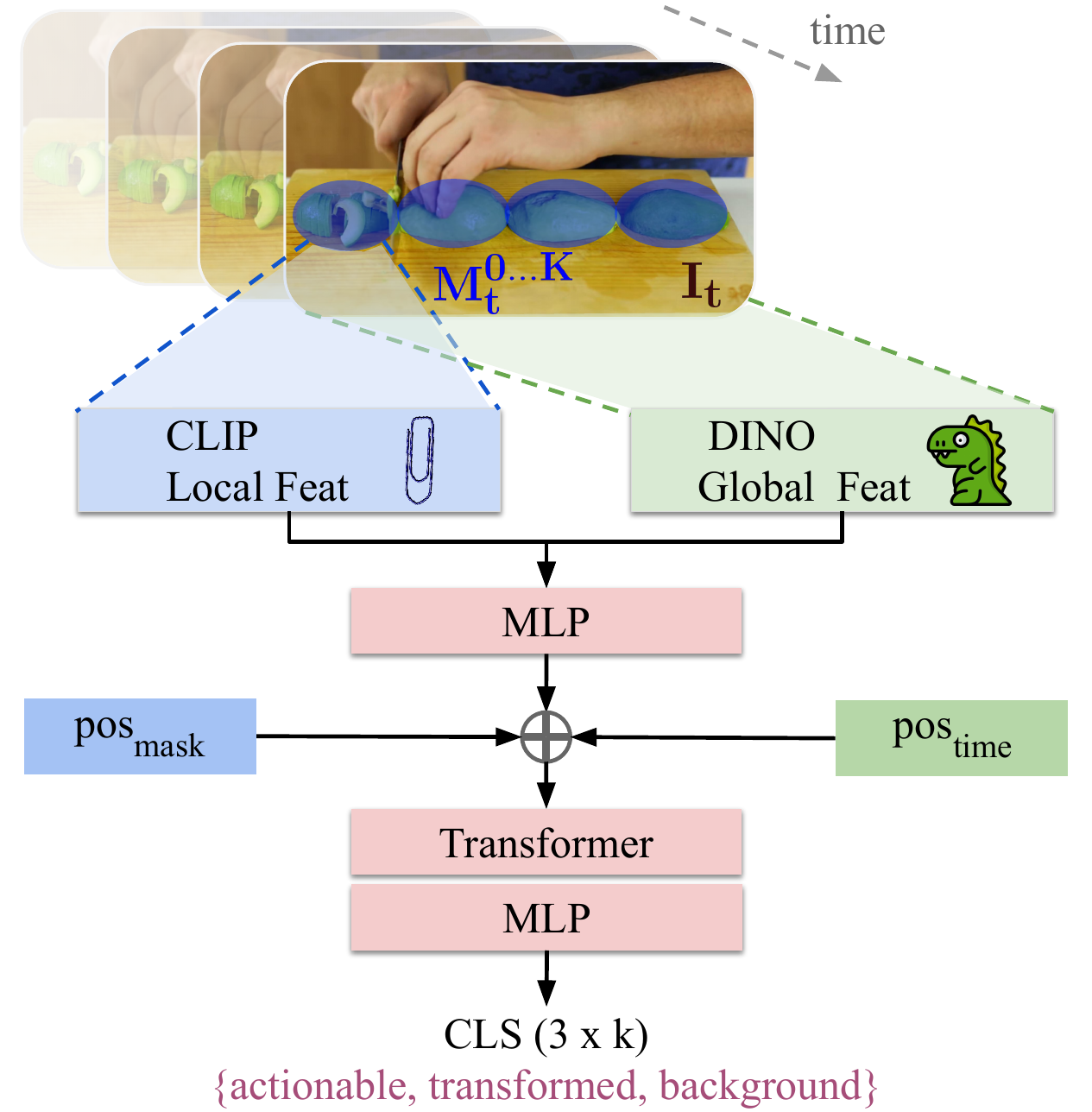}
    \caption{
    Model architecture (Sec.~\ref{subsec:model_training})
    }
    \label{fig:arch}
\vspace{-1.5em}
\end{figure}

We now describe the components involved in training our model 
on the generated pseudo-labels.
We train a separate model for each verb (action), following prior work on OSC classification~\cite{souvcek2022look,xue2023vidosc}. 
Generalizing to unseen actions would entail training an action-conditioned multi-task model, which we leave to future work.
The architecture (Fig.~\ref{fig:overview}c, Fig.~\ref{fig:arch}) consists of a video encoder and a transformer-based decoder that predicts state labels for each mask proposal. 
Given video frames \(\{I_0, I_t, \ldots, I_T\}\) and object masks \(\{M_t^{0..K}\}\), we extract \textit{local features} from CLIP~\citep{clip} and \textit{global features} from DINOv2~\citep{oquab2023dinov2}. CLIP provides fine-grained mask-level cues, while DINO supplies frame-level context. We aggregate these features, add mask- and time-positional embeddings \((\text{pos}_{\text{mask}}, \text{pos}_{\text{time}})\), and process them with a transformer that models temporal dynamics. A cross-entropy loss trains the model to classify each region as \{actionable, transformed, background\}.  
By combining CLIP and DINO features with temporal modeling, the model captures both detailed object states and their progression across frames. 
Note that while our baseline classifies segment proposals, the SPOC task on the WTC dataset is fundamentally a pixel-level segmentation benchmark.

\section{The \dataset\ Dataset}
\label{sec:dataset}

Existing OSC datasets do not capture the fine-grained 
information necessary to  evaluate segmenting \task OSCs. To address this gap, we introduce \dataset, a large-scale dataset featuring detailed intra-object state-change annotations across a wide variety of objects and actions. Below, we provide details on data collection and annotation with full details in Appendix~\ref{sec_supp:dataset}.

\custompar{Data Collection}
We source our dataset from the 
recent \htc\ dataset~\cite{xue2023vidosc}.
Derived from HowTo100M~\cite{miech2019howto100m}, which comprises a vast collection of YouTube videos, \htc\ provides frame-level state labels for cooking activity videos.
In building \dataset\ (WTC for short), we take take all OSCs from \htc\ that clearly exhibit gradual state change across object regions (e.g., chopping, peeling, mashing), while excluding those where objects change uniformly all at once (e.g., blending, grilling), resulting in 10 OSCs
(see Table~\ref{tab:htc_vost_big}).
Using the pseudo-labeling process (Sec.~\ref{subsec:pseudo_label_gen}), we construct a training set
of
\stat{$\sim$17k}
videos, encompassing 
\stat{$\sim$700k}
frames, with an average clip duration of \stat{41} seconds. \textbf{All evaluation sets are manually labeled} (see Sec.~\ref{sec:setup}).
\dataset\ includes \stat{153} OSCs spanning \stat{73} objects labeled as seen across train and test, and \stat{60} OSCs spanning \stat{43} objects labeled as novel during testing.

\custompar{Dataset Comparison}
Table~\ref{tab:dataset} compares our dataset with prior OSC video benchmarks. \dataset\ is the first to target \task segmentation, providing fine-grained intra-object state-change labels absent from earlier work. Among segmentation datasets, it offers unmatched scale—\stat{9}$\times$ more training data and \stat{8}$\times$ more evaluation data than existing benchmarks~\cite{tokmakov2023vost,Yu_2023_vscos}. This expansion delivers both higher granularity and greater utility of OSC annotations, setting a new standard for the field.

\begin{table}[!t]
\tablestyle{1.1pt}{1.1}
  \centering
  \begin{tabular}{lccccccc}
    \toprule
    Datasets & \# Obj & \# ST & \# OSC & \makecell[c]{\# Videos\\ (train / eval)} & Seg & \makecell[c]{OSC\\ Label} & \makecell[c]{Intra-\\ object} \\
    \midrule
    ChangeIt~\cite{souvcek2022look} & 42 & 27 & 44 & 34,428 / 667\hspace{0.35em} & \redcross & \greencheck & \redcross \\
    \htc~\cite{xue2023vidosc} & 134 & 20 & 409 & 34,550 / 5,424 & \redcross & \greencheck & \redcross \\
    \addlinespace \hdashline \addlinespace
    VSCOS~\cite{Yu_2023_vscos} & 124 & 30 & 271 & 1,905 / 98\hspace{0.75em} & \greencheck & \redcross & \redcross \\
    VOST~\cite{tokmakov2023vost} & 57 & 25 & 51 & \hspace{0.25em}713 / 141 & \greencheck & \redcross & \redcross \\
    \rowcolor{LighterGray}
    \dataset ~(ours) & 116 & 10 & 213 & \stat{16,999} / \stat{1,162}\hspace{0.15em} & \greencheck & \greencheck & \greencheck \\
    \bottomrule
  \end{tabular}
  \caption{\textbf{Comparison with existing video OSC datasets.} `Obj' and `ST' represent objects and state transitions, respectively. We present the first benchmark for \task video OSC, with fine-grained intra-object state-change segmentations. Among segmentation datasets, we offer unparalleled scale, while significantly enhancing state-change granularity.
  } 
  \label{tab:dataset}
    \vspace{-1.5em}
\end{table}

\section{Experiments}
\label{sec:expts}

We define datasets, metrics, baselines then present results.

\subsection{Setup}\label{sec:setup}

\begin{table*}[h]
\tablestyle{4pt}{1.15}
\centering
\aboverulesep=0ex
\belowrulesep=0ex
\begin{NiceTabular}{l|c|cccccccccc|c|cc}
    \toprule
    \textbf{Model} & \multicolumn{11}{c}{\textbf{WTC-HowTo}} & \multicolumn{3}{c}{\textbf{WTC-VOST}} \\
     & \textbf{\underline{Mean}} & \textbf{Chop} & \textbf{Crush} & \textbf{Coat} & \textbf{Grate} & \textbf{Mash} & \textbf{Melt} & \textbf{Mince} & \textbf{Peel} & \textbf{Shred} & \textbf{Slice} & \textbf{\underline{Mean}} & \textbf{Chop} & \textbf{Peel} \\
    \midrule
    \multicolumn{15}{c}{\textbf{\textit{w/o training}}} \\
    MaskCLIP~\cite{zhou2022maskclip} & 0.146 & 0.165 & 0.095 & 0.134 & 0.173 & 0.163 & 0.107 & 0.110 & 0.154 & 0.206 & 0.153 & 0.098 & 0.111 & 0.085 \\
    GroundedSAM~\cite{ren2024gsam} & 0.273 & 0.286 & 0.235 & 0.275 & 0.317 & 0.272 & 0.173 & 0.318 & 0.256 & 0.310 & 0.284 & 0.221 & 0.271 & 0.172 \\
    SAM-Track~\cite{cheng2023samtrack} & 0.275 & 0.289 & 0.216 & 0.323 & 0.276 & 0.311 & 0.198 & 0.286 & 0.283 & 0.294 & 0.272 & 0.194 & 0.227 & 0.161 \\
    Random Label & 0.274 & 0.278 & 0.242 & 0.279 & 0.269 & 0.303 & 0.275 & 0.272 & 0.259 & 0.295 & 0.269 & 0.184 & 0.190 & 0.179 \\
    DEVA~\cite{cheng2023deva} & 0.282 & 0.321 & 0.198 & 0.256 & 0.333 & 0.289 & 0.223 & 0.328 & 0.211 & 0.370 & 0.293 & 0.228 & 0.298 & 0.157 \\
    SAM+GPT-4o~\cite{openai2023gpt4} & 0.427 & 0.401 & 0.379 & 0.391 & 0.458 & 0.475 & \textbf{0.470} & 0.456 & 0.389 & 0.470 & 0.382 & 0.189 & 0.201 &  0.176  \\
    \rowcolor{LighterGray}
    SPOC\ (PL) & 0.411 & 0.438 & 0.351 & 0.237 & 0.429 & 0.527 & 0.416 & 0.493 & 0.373 & 0.418 & 0.428 & 0.204 & 0.229 & 0.179 \\
    \rowcolor{LighterGray}
    \quad +CO  & 0.438 & 0.456 & 0.371 & 0.378 & 0.441 & 0.553 & 0.415 & 0.497 & 0.380 & 0.442 & 0.445  & 0.216 & 0.241 & 0.191 \\
    \rowcolor{LighterGray}
    \quad +AR  & 0.455 & 0.461 & 0.383 & 0.447 & 0.447 & 0.566 & 0.418 & 0.500 & 0.417 & 0.453 & 0.458  & 0.229 & 0.257 & 0.201 \\
    \midrule
    \multicolumn{15}{c}{\textbf{\textit{w/ training}}} \\
    MaskCLIP+~\cite{zhou2022maskclip} & 0.182 & 0.200 & 0.125 & 0.168 & 0.216 & 0.197 & 0.138 & 0.143 & 0.194 & 0.256 & 0.186 & 0.119 & 0.136 & 0.103 \\
    GroundedSAM~\cite{ren2024gsam} & 0.275 & 0.291 & 0.225 & 0.282 & 0.323 & 0.274 & 0.171 & 0.325 & 0.259 & 0.313 & 0.285 & 0.223 & 0.275 & 0.171 \\
    \rowcolor{LighterGray}
    SPOC & \textbf{0.502} &  \textbf{0.523}	 & \textbf{0.422}	 & \textbf{0.526}	 & \textbf{0.528}	 & \textbf{0.610}	 & 0.425 & \textbf{0.541}	 & \textbf{0.449}	 & \textbf{0.503}	 & \textbf{0.494} & \textbf{0.267} & \textbf{0.307} & \textbf{0.227} \\
    \bottomrule
\end{NiceTabular}
\caption{\textbf{Results on WhereToChange-HowTo and WhereToChange-VOST.} 
Comparison of different models based on \cc{Jaccard score }mIoU performance. Our pseudo-labeling strategy (Sec.~\ref{subsec:pseudo_label_gen}: PL=pseudo-labels) substantially outperforms existing state-of-the-art object segmentation baselines, with dynamics constraints (Sec.~\ref{subsec:constraints}: CO=causal-ordering, AR=ambiguity-resolution) further enhancing performance. The fully trained model (Sec.~\ref{subsec:model_training}) achieves the highest accuracy, benefiting from large-scale training. Notably, our model trained on HowTo also
outperforms the baselines on VOST, a challenging out-of-distribution dataset comprising continuously captured egocentric videos.
}
\label{tab:htc_vost_big}
\vspace{-1em}
\end{table*}

\custompar{Datasets} 
For evaluation, we obtain ground truth human annotations for video clips from two datasets: WTC-HowTo
and WTC-VOST. 
While the former is sourced from large-scale YouTube instructional videos~\cite{miech2019howto100m} of free-form, mostly third-person videos with diverse objects, the latter~\cite{tokmakov2023vost} consists of continuously captured human activity from egocentric video datasets Ego4D~\cite{grauman2022ego4d} and EPIC-Kitchens~\cite{Damen2018EPICKITCHENS}. 
Within VOST, we consider two \task OSC categories: chop and peel. 
The combination of datasets provides a comprehensive testbed,
and allows testing our HowTo-trained model on a distinct dataset.

Professional annotators manually annotate \stat{1001} and \stat{155} videos from \htc~and VOST to form our evaluation sets WTC-HowTo and WTC-VOST, respectively.
Each annotator is presented with frames from a video clip and the corresponding OSC category. Their task is to 
label pixels on
the object of interest in each frame as either actionable or transformed. Clips with ambiguity in distinguishing between states are identified and excluded from the evaluation set, ensuring high quality. The annotation process was conducted on TORAS~\cite{toras} with the assistance of \stat{7} human annotators, totaling 
\stat{~350} hours.\footnote{Our training uses only pseudo-labels and does not require ground truth labels. The annotations are reserved exclusively for evaluation.}

\custompar{Metrics and Data Splits}
We report Jaccard Index ($\mathcal{J}$), also called Intersection over Union (IoU), between predicted and
ground truth masks, computed as the mean IoU over actionable and transformed masks.
Following~\cite{tokmakov2023vost,shen2024rosa}, we do not report a boundary $\mathcal{F}$-measure since object contours in OSCs are ill-defined due to drastic transformations over time.

In addition to the full test sets, we provide results on three distinct subsets of \dataset:
WTC-Transition, focusing exclusively on transition frames from \htc\ (\stat{8405} frames, or \stat{23}\%), and WTC-Seen/Novel, consisting of seen/novel object categories (novel = unseen during training).  WTC-Transition isolates moments where the object undergoes significant state changes (excluding frames in the initial or end state), emphasizing the progressive spatial evolution of the object.
 WTC-Novel isolates the generalization ability to novel objects (e.g., seeing an apple being chopped at test time after only seeing other objects be chopped during training).

\custompar{Implementation}
We sample videos at 1 fps for both training and annotation collection. 
We apply GroundingDino~\cite{liu2023gdino} (for object detection) and SAM~\cite{kirillov2023sam} (for mask proposal generation) every 10 frames and track masks using DeAOT~\cite{yang2022deaot} at intermediate frames. 
Using ViT-B/32 as our CLIP~\cite{clip} vision encoder for pseudo-labels, we set similarity thresholds in Eq.~\ref{eq:clip_thr} using grid search.
We train a separate model for each verb following prior work on OSC classification~\cite{souvcek2022look,xue2023vidosc}.
Full details in Appendix~\ref{subsec_supp:implementation}.

\begin{figure*}[t]
    \centering
    \includegraphics[width=0.95\linewidth]{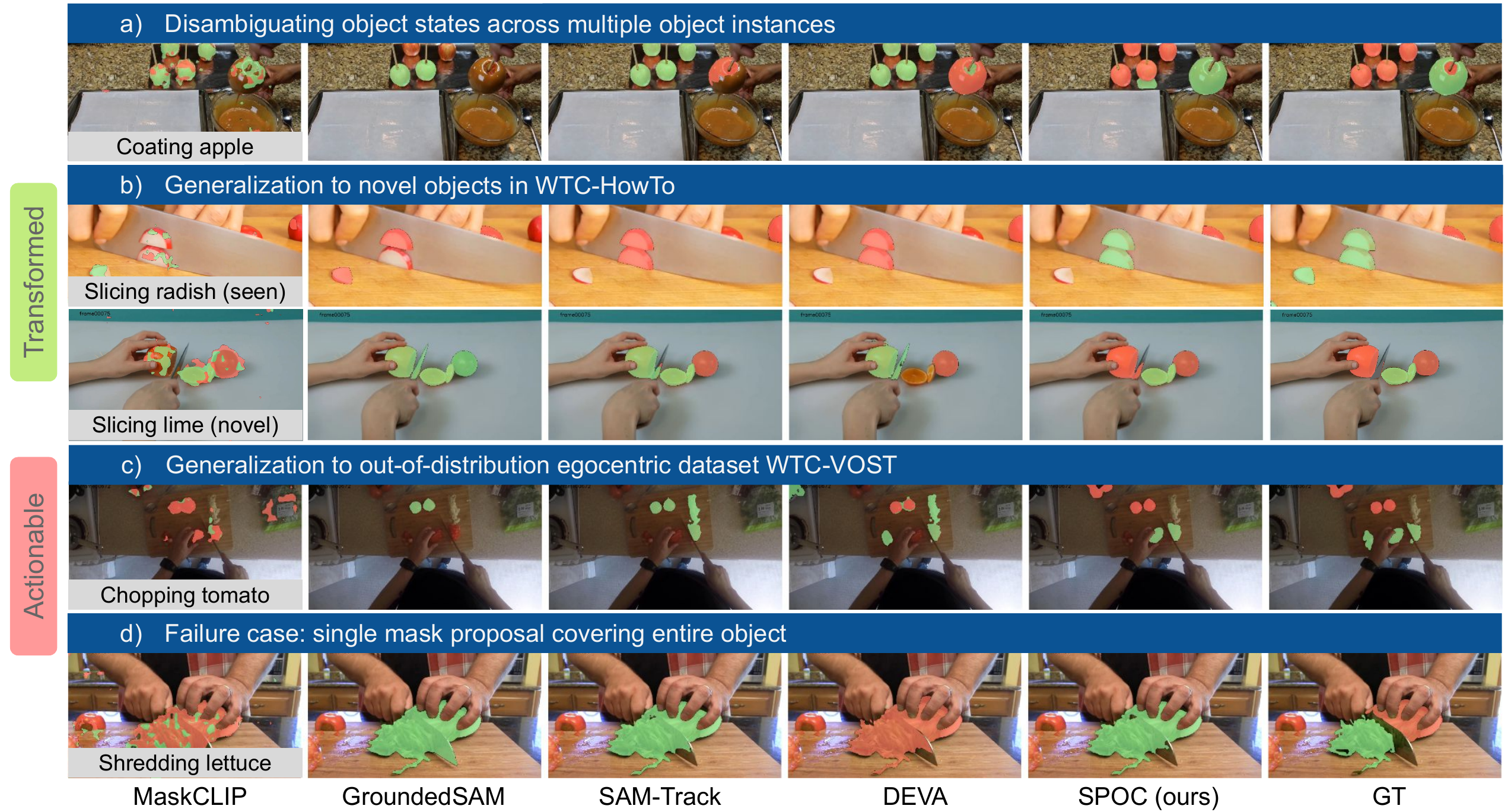}
    \caption{\textbf{Qualitative results. }
    a) SPOC clearly distinguishes between actionable and transformed instances of the state-changing object (coated  vs uncoated apple),  b) with the ability to generalize to novel unseen objects (slicing lime). In contrast, baseline methods tend to be state-change agnostic with decreased ability to disambiguate object states. c) SPOC also shows good generalization to the challenging out-of-distribution VOST dataset. d) Failure cases arise from singular mask proposals spanning the entire object during transitions (single mask for the full lettuce), affecting the model's intra-object segmentation capability.
    }
    \label{fig:main_results}
    \vspace{-1em}
\end{figure*}

\subsection{Baselines}
Since no prior methods tackle the proposed task, we 
adapt existing high-quality open-vocabulary segmentation works for our task.
They broadly fall under two categories:
\custompar{(1)   Pixel-based}: Classify each pixel in the image into one of 3 categories---actionable, transformed, or background.
\begin{itemize}[leftmargin=1em]
    \item[--] \textbf{MaskCLIP}~\cite{zhou2022maskclip}: predicts dense segmentation by minimally adapting the final layers of CLIP.
    \item[--] \textbf{MaskCLIP+}~\cite{zhou2022maskclip}: finetunes DeepLab-v2~\cite{chen2017deeplab} on the pseudo-labels generated by MaskCLIP on WTC-HowTo.
\end{itemize}
\custompar{(2)   Object-centric}: Use open-vocabulary object detectors and segmentors to obtain relevant object regions and then classify them into the above 3 categories.
\begin{itemize}[leftmargin=1em]
    \item[--] \textbf{GroundedSAM}~\cite{ren2024gsam}: uses GroundingDino~\cite{liu2023gdino} to obtain text-prompted bounding boxes, which are then used to prompt SAM~\cite{kirillov2023sam} for masks.
    \item[--] \textbf{SAM-Track}~\cite{cheng2023samtrack}: incorporates DeAOT~\cite{yang2022deaot} into GroundedSAM~\cite{ren2024gsam} to propagate masks through time.
    \item[--] \textbf{Random Label}: a naive baseline that randomly assigns actionable or transformed labels to SAM-Track masks.
    \item[--] \textbf{DEVA}~\cite{cheng2023deva}: a decoupled video segmentation approach that uses XMem~\cite{cheng2022xmem} to store and propagate masks.
\end{itemize}
\custompar{(3)   Part-centric}: Variant of object-centric methods where intra-object parts are classified by a VLM.
\begin{itemize}[leftmargin=1em]
    \item[--] \textbf{SAM}~\cite{ravi2024sam2} + \textbf{GPT-4o}~\cite{openai2023gpt4}: use \cite{ren2024gsam} for object segmentation, extract intra-object parts via a point-grid prompt to ~\cite{ravi2024sam2}, and query GPT-4o to classify each part into one of the above 3 categories. This tests whether state-of-the-art VLMs can provide fine-grained, state-sensitive labels.
\end{itemize}
Through the above baselines, we aim to comprehensively benchmark prior state-of-the-art open-vocabulary video object segmentation methods and VLMs on our novel task.
We finetune baselines on WTC-HowTo where feasible (MaskCLIP+, GroundedSAM).
Details in Appendix~\ref{subsec_supp:baselines}.

\subsection{Results}

\begin{table}[t]
\tablestyle{3.4pt}{1.1}
\centering
\begin{tabular}{lc|cccc}
    \toprule
    \textbf{Model} & \textbf{w/o OSC}  & \textbf{Full}  &  \textbf{Transition} &  \textbf{Seen}  &  \textbf{Novel} \\
    \midrule
    MaskCLIP~\cite{zhou2022maskclip} & 0.346 & 0.146 & 0.152 & 0.148 & 0.144\\
    MaskCLIP+~\cite{zhou2022maskclip} & 0.393 & 0.182 & 0.181 & 0.188 & 0.184\\
    GroundedSAM~\cite{ren2024gsam} & 0.611 & 0.273 & 0.262 & 0.276 & 0.270 \\
    SAM-Track~\cite{cheng2023samtrack} & 0.534 & 0.275 & 0.240 & 0.279 & 0.270 \\
    DEVA~\cite{cheng2023deva} & 0.519 & 0.282 & 0.269 & 0.283 & 0.281 \\
    \rowcolor{LighterGray}
    SPOC (ours) & 0.555 & \textbf{0.502} & \textbf{0.332} & \textbf{0.523} & \textbf{0.492} \\
    \bottomrule
\end{tabular}
\caption{\textbf{Performance across different data splits of WTC-HowTo. } Owing to large-scale training, SPOC shows strong generalization to novel objects. 
} 
\label{tab:htc_abl}
\vspace{-2em}
\end{table}

\begin{figure*}[t!]
    \centering
    \includegraphics[width=0.95\linewidth]{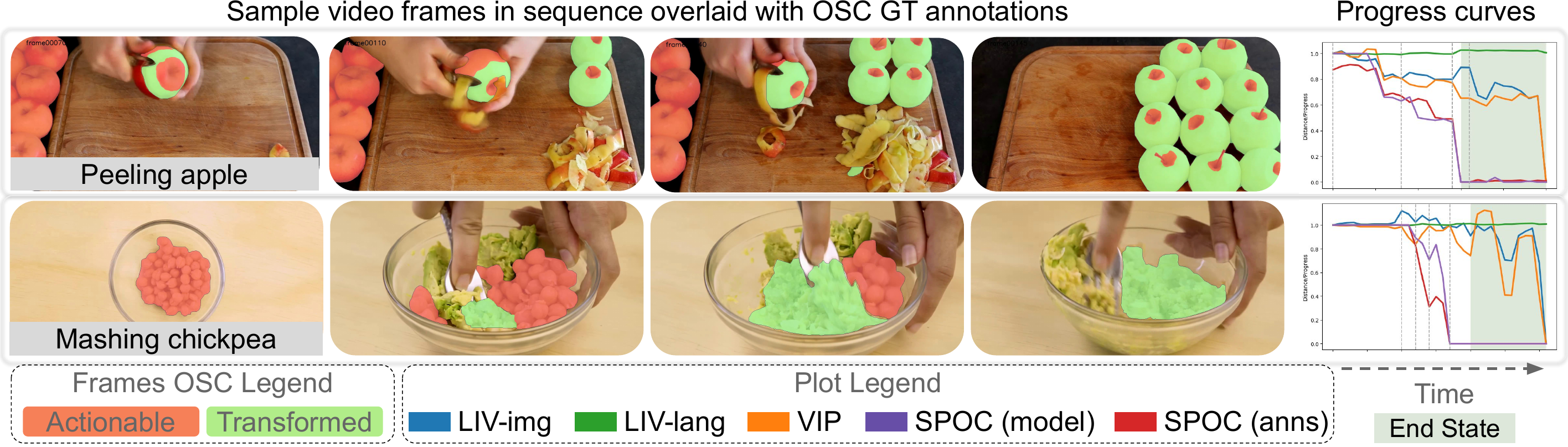}
    \caption{\textbf{Activity progress curves. }
    We show sample frames from a video sequence with progress curves generated by different methods, where vertical lines indicate the time-steps of sampled frames. 
    Ideal curves should decrease monotonically, and saturate upon reaching the end state. 
    In contrast to goal-based representation learning methods such as VIP~\cite{ma2022vip} and LIV~\cite{ma2023liv}, OSC-based curves accurately track task progress, making them valuable for downstream applications like progress monitoring and robot learning.}
    \label{fig:progress}
    \vspace{-1em}
\end{figure*}

In our experiments, we aim to answer \edit{six} key questions:

\custompar{Can \method\ segment \task OSCs?}
Table~\ref{tab:htc_vost_big} shows SPOC outperforms all baselines.
Among existing approaches, object-centric methods outperform pixel-based ones, which tend to introduce higher noise in dense predictions. However, their performance remains close to the naive Random Label baseline, highlighting the difficulty of fine-grained state distinction.
Our pseudo-labeling strategy alone (SPOC (PL)) already surpasses all segmentation baselines, and training our model further improves performance (SPOC, w/ training). 
From qualitative results shown in Fig.~\ref{fig:main_results}, our method clearly distinguishes between actionable and transformed instances of the same object. E.g., for coating apple, where multiple apples transition between uncoated and coated states, SPOC accurately classifies each apple by its state. 
In contrast, baseline methods—even when fine-tuned with our training videos—remain largely state-agnostic and struggle to disambiguate between the two states.
Importantly, the part-centric baseline with GPT-4o—a state-of-the-art VLM—also fails to match SPOC, underscoring that the proposed benchmark is far from solved and presents a real challenge for current VLM and VOS models.
Failure cases (Fig.~\ref{fig:main_results}d) typically arise when single mask proposals cover the entire object during transitions, limiting the model's effectiveness in achieving fine-grained intra-object segmentation (Table~\ref{tab:htc_abl}-Transition).
Baseline performance is explained in Appendix~\ref{subsec_supp:baselines},
while additional ablations of SPOC are provided in Appendix~\ref{sec_supp:ablations}.

\custompar{Is spatially segmenting OSCs more challenging than segmenting the object itself?}
To further evaluate the complexity of this task, we assess performance in predicting the full object segmentations on the same test videos, i.e., by fusing the actionable/transformed masks into a single, state-agnostic object mask (Table~\ref{tab:htc_abl}-w/o OSC). Notably, these results are considerably better than those for fine-grained classification into ‘actionable’ and ‘transformed’ states, 
showing that existing models struggle with fine-grained state distinction, \emph{even if they can capture the object relatively well}.

\custompar{Does \method\ generalize to novel OSCs?}
We retain the seen/novel object splits from \htc\ and evaluate generalization to novel objects on WTC-HowTo (Table~\ref{tab:htc_abl}). 
Our large-scale training enables strong generalization, highlighting the role of high-quality pseudo-labels in enhancing performance.  E.g., the model learns concepts like “sliced pieces” during training, allowing it to segment novel objects into whole and sliced pieces during testing (Fig.~\ref{fig:main_results}b).

\custompar{Does \method\ generalize to new datasets?}
To illustrate robustness under domain shift,
we also perform cross-dataset evaluation on WTC-VOST, a challenging dataset of out-of-distribution, continuously captured egocentric videos.
From Table~\ref{tab:htc_vost_big}, the SPOC model trained on HowTo achieves the best overall performance 
(Fig.~\ref{fig:main_results}c). 
However, a primary limitation in this dataset was unreliable mask tracking of SAM-Track~\cite{cheng2023samtrack}, the off-the-shelf tracker we employed, which struggled with the sudden, jerky head movements common in egocentric video. Future advancements in object detection and mask tracking are likely to further enhance our model's performance on these complex video sequences.

\custompar{How important are the state-change dynamics constraints?}
To test the efficacy of our state-change dynamics constraints (Sec.~\ref{subsec:constraints}), we progressively incorporate each component and evaluate performance in Table~\ref{tab:htc_vost_big}. Starting with the CLIP-similarity-based pseudo-labels, each additional dynamics constraint markedly improves segmentation, culminating in an overall improvement of 22\%.

\begin{table}[t]
\tablestyle{4pt}{1.1}
\centering
\begin{tabular}{lcccc}
    \toprule
    \textbf{Model} & \multicolumn{3}{c}{\hspace{-0.6em}\textbf{WTC-HowTo}} & \multicolumn{1}{c}{\hspace{-0.6em}\textbf{WTC-VOST}} \\
    & $\tau\downarrow$  &  $end_\sigma\downarrow$ &  $end_{l2}\downarrow$  &  $\tau\downarrow$\\
    \midrule
    VIP~\cite{ma2022vip} & -0.437 & 0.167 & 0.639 &-0.405 \\
    LIV-lang~\cite{ma2023liv} & -0.031 &  - & 0.999 & -0.062\\
    LIV-lang (finetune)~\cite{ma2023liv} & 0.054 & - & 1.005 &0.048\\
    LIV-img~\cite{ma2023liv} & -0.468 & 0.114& 0.644 & -0.401\\
    LIV-img (finetune)~\cite{ma2023liv} & -0.418& 0.113 &  0.673 &-0.374\\
    \rowcolor{LighterGray}
    SPOC (model)  & \textbf{-0.611} & \textbf{0.057} & \textbf{0.212} & \textbf{-0.581} \\
    \rowcolor{DarkerGray}
    SPOC (annotations) & -0.670 & 0.031 & 0.206 & -0.651\\
    \bottomrule
\end{tabular}
\caption{\textbf{Activity progress metrics.} OSC-based curves outperform state-of-the-art representation learning methods for robot-learning. $\tau$ represents Kendall's Tau as the monotonicity index; $end_\sigma$ and $end_{l2}$ are task completion metrics.}
\label{tab:progress}
\vspace{-1.5em}
\end{table}

\custompar{How could intra-object \task OSCs be used for robotics?}
A key motivation for this task is activity progress monitoring, crucial for AR/MR and robotics. We test this by applying SPOC outputs as a continuous progress reward for robot learning, defined as  
$R_t=\frac{A_t^{act}}{A_t^{act} \cup A_t^{trf}}$,
where $A_t^{act}$ and $A_t^{trf}$ are the actionable and transformed areas at time $t$.  
We compare against LIV~\cite{ma2023liv} and VIP~\cite{ma2022vip}, state-of-the-art methods that learn goal-based representations from human-activity videos~\cite{grauman2022ego4d,Damen2018EPICKITCHENS}. As shown in Fig.~\ref{fig:progress}, ideal reward curves decrease monotonically and saturate at task completion~\cite{ma2023liv,ma2022vip}. SPOC-generated curves closely follow this behavior, accurately reflecting task progress.  
Quantitative metrics on monotonicity and completion (Table~\ref{tab:progress}, details in Appendix~\ref{sec_supp:progress}) 
show that SPOC significantly outperforms baselines. These results highlight the downstream utility of fine-grained OSC segmentation: providing real-time, progress-aware feedback for robotic learning.

\vspace{-0.5em}
\section{Conclusion}
\label{sec:concl}

We introduce spatially-progressing OSC segmentation, a novel task to track how object state changes unfold at the pixel level. To support it, we build WhereToChange, the first large-scale benchmark with fine-grained intra-object state-change annotations across diverse activities.
Using VLM-guided pseudo-labeling with dynamics constraints, we offer a baseline that surpasses existing methods and generalizes to novel objects. Yet, results show that even state-of-the-art VLMs and segmentation models struggle on this task, leaving the benchmark far from solved.
Our work highlights this gap and positions spatial OSC segmentation as a benchmark challenge for spatio-temporal reasoning about objects, with direct implications for AR/MR activity tracking and robotic learning.


\section*{Acknowledgements}
\PM{We extend our gratitude to TORAS~\cite{toras} for providing the segmentation annotation platform, with special thanks to Tianxing Li for his prompt assistance and support.}

{
    \small
    \bibliographystyle{ieeenat_fullname}
    \bibliography{main}
}

\clearpage

\counterwithin{section}{part} 

\setcounter{section}{0}
\setcounter{figure}{0}
\setcounter{table}{0}
\renewcommand{\thesection}{\Alph{section}}
\renewcommand{\thetable}{\Alph{table}}
\renewcommand{\thefigure}{\Alph{figure}}

\newif\ifarxiv
\arxivtrue

\ifarxiv
    \section*{\Large Appendix}
\else
    \maketitlesupplementary
\fi

The supplementary materials consist of:
\begin{itemize}
\item (\ref{sec_supp:video}) A supplementary video offering a comprehensive overview of \method\ along with qualitative examples.
\item (\ref{sec_supp:dataset}) Detailed information on \dataset, elaborating on the data collection and annotation process. In addition, we analyze the dataset along various axes, highlighting its wide-ranging and expansive nature.  
\item (\ref{sec_supp:expt_setup}) Experimental details: this includes complete experimental setup, baselines and metrics.
\item (\ref{sec_supp:ablations}) Extensive ablations of various component of SPOC with further results and visualizations that are referenced in the main paper.
\item (\ref{sec_supp:progress}) Details of the downstream task on activity progress including metrics and baselines.
\item (\ref{sec_supp:limitations}) Discussion on limitations and scope for future work.
\end{itemize}

\section{Video Containing Qualitative Results}
\label{sec_supp:video}
We invite the reader to view the video available at \url{https://vision.cs.utexas.edu/projects/spoc-spatially-progressing-osc}, where we provide: (1) a comprehensive overview of \method, (2) video examples from \dataset, and (3) qualitative examples of \method's predictions. These examples highlight \method's ability in disambiguating object states across multiple object instances and effectively distinguishing actionable and transformed object regions. It delivers temporally smooth and coherent predictions that follow the natural OSC causal and temporal dynamics (from actionable to transformed), and shows strong performance even with novel OSCs not seen in training. Moreover, the activity progress curves highlight the importance of our proposed \task\ task for downstream progress-monitoring and robotics applications. All these underscore the efficacy of \method, our proposed \dataset\ dataset, and and our novel \task task.

\section{The \dataset\ Dataset} 
\label{sec_supp:dataset}
In this section, we describe in detail our annotation pipeline (Sec.~\ref{subsec_supp:annotation}), analyze various properties of our \dataset\ dataset (Sec.~\ref{subsec_supp:dataset_analysis}), and provide diverse samples from our dataset spanning a wide-range of state-changes and objects (Sec.~\ref{subsec_supp:dataset_samples}).

\subsection{OSC Taxonomy}

\begin{table*}[!t]
\tablestyle{5pt}{1.4}
  \centering
\begin{tabular}{lp{10cm}p{5cm}}
\toprule
Verb & Objects (seen) & Objects (novel) \\
\hline
\rowcolor{LighterGray}
\multicolumn{3}{c}{\hspace{-0.6em}\textbf{WTC-HowTo}} \\
\hline
chopping &
apple, avocado, bacon, banana, basil, broccoli, cabbage, carrot, celery, chicken, chili, chive, chocolate, cucumber, egg, garlic, ginger, jalapeno, leaf, lettuce, mango, mushroom, nut, onion, peanut, pecan, pepper, potato, scallion, shallot, strawberry, tomato, zucchini &
almond, butter, capsicum, cauliflower, chilies, date, leek, pineapple, sausage
\\
coating &
apple, bread, cake &

\\
crushing &
biscuit, cooky, garlic, ginger, peanut, potato, strawberry, tomato &
pepper
\\
grating &
apple, carrot, cheese, chocolate, cucumber, lemon, onion, orange, parmesan, potato, zucchini &
butter, cauliflower, coconut, mozzarella
\\
mashing &
avocado, banana, chickpea, garlic, potato, strawberry, tomato &
egg
\\
melting &
butter, candy, caramel, gelatin, ghee, jaggery, margarine, marshmallow, shortening, sugar &

\\
mincing &
beef, cilantro, garlic, ginger, jalapeno, meat, onion, parsley, scallion, shallot &
carrot, pepper, tomato
\\
peeling &
apple, avocado, banana, beet, carrot, cucumber, egg, eggplant, garlic, ginger, lemon, mango, onion, orange, pear, plantain, potato, pumpkin, shrimp, squash, tomato, zucchini &
kiwi, peach, pineapple, shallot
\\
shredding &
beef, cabbage, carrot, cheese, chicken, lettuce, meat, pork, potato, zucchini &
coconut, mozzarella, parmesan
\\
slicing &
apple, avocado, bacon, banana, beef, bread, cabbage, cake, carrot, chicken, cucumber, egg, eggplant, ginger, leek, lemon, mango, meat, mushroom, onion, peach, pear, pepper, pineapple, potato, radish, sausage, scallion, shallot, steak, strawberry, tofu, tomato, watermelon, zucchini &
butter, celery, jalapeno, lime, mozzarella, olive, orange, pepperoni
\\
\hline
\rowcolor{LighterGray}
\multicolumn{3}{c}{\hspace{-0.6em}\textbf{WTC-VOST}} \\
\hline
chopping &bacon, broccoli, carrot, celery, chicken, chili, cucumber, garlic, ginger, mango, onion, pepper, potato, scallion, spinach, tomato
 & asparagus, aubergine, beef, bread, butter, cake, corn, courgette, dough, gourd, ham, herbs, ladyfinger, mozzarella, pea, peach, pumpkin, salad
\\
peeling & banana, carrot, garlic, onion, potato
 & aubergine, courgette, fish, gourd, peach, root
 \\
\bottomrule
\end{tabular}
  \caption{\textbf{OSC taxonomy for \dataset. }WTC encompasses \stat{116} objects undergoing \stat{10} distinct state transitions, resulting in \stat{232} unique OSCs (\stat{170} seen and \stat{62} novel) across both WTC-HowTo and WTC-VOST.}
  \label{tab:osc_taxonomy}
\end{table*}

Our primary dataset, WTC-HowTo, is a large-scale collection derived from HowToChange~\cite{xue2023vidosc}, focusing on 10 state-change verbs that exhibit spatial progression behavior. These include actions such as chopping, coating, and mashing, where object regions undergo sequential transformations from one state to another. From the 20 verbs in HowToChange, we select the 10 that demonstrate this characteristic, excluding verbs like blending, frying, and browning, which typically transform objects uniformly without intra-object state distinctions.
The complete OSC object taxonomy is provided in Table~\ref{tab:osc_taxonomy}. Novel objects are absent for three verbs (coat, crush, melt) in WTC-HowTo, hence only seen objects are reported for these actions. Overall, our dataset encompasses a diverse range of objects across both subsets, providing a comprehensive benchmark for state-change understanding.

\subsection{Annotation Pipeline for Eval Set}
\label{subsec_supp:annotation}

\begin{figure*}[htpb]
    \centering
    \includegraphics[width=\linewidth]{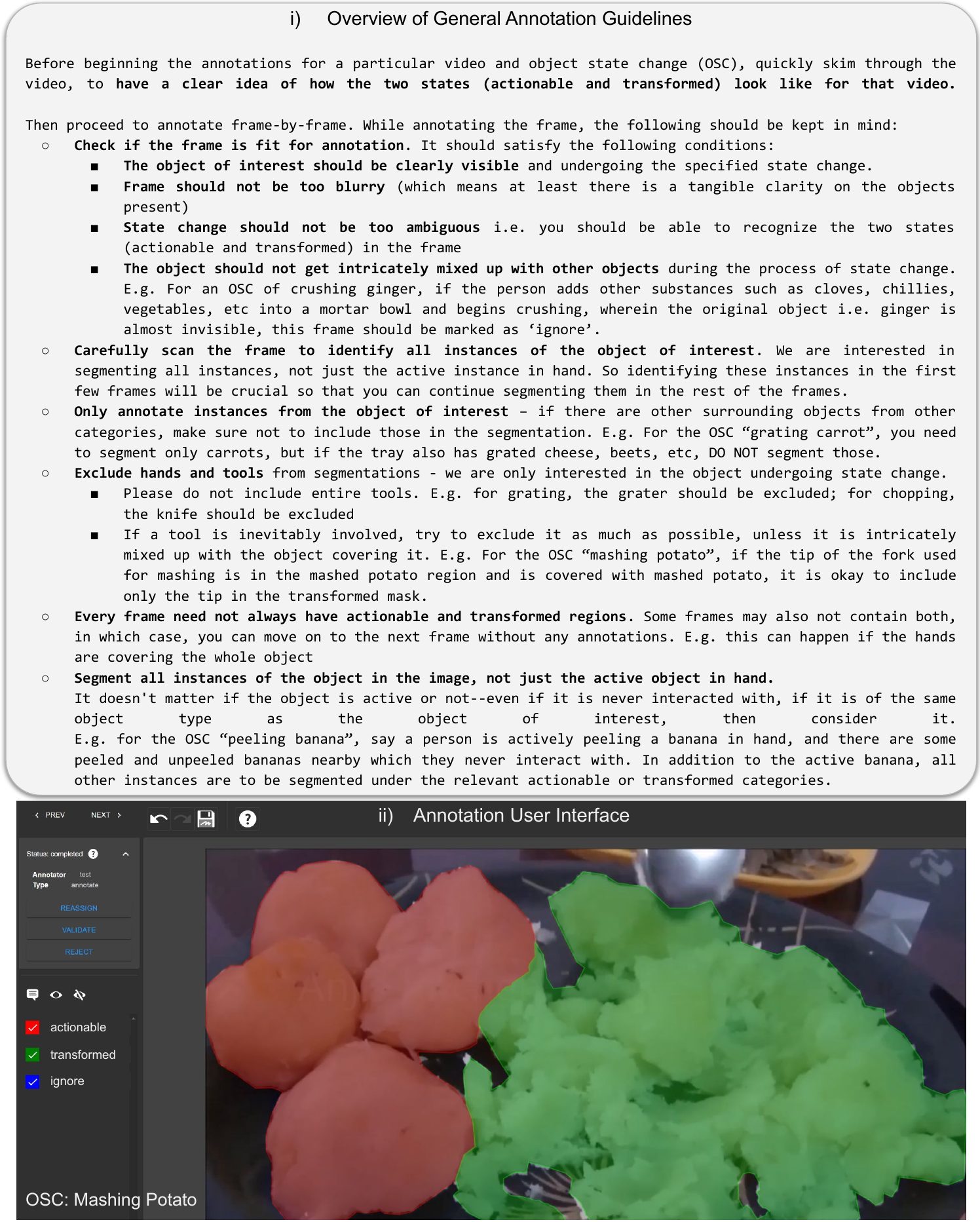}
    \caption{\textbf{Annotation User Interface and Guidelines. }
    \textbf{(i)} Overview of general annotation guidelines provided to the annotators. Verb-specific guidelines are in Table~\ref{tab:ann_guidelines}.
    \textbf{(ii)} Annotation user interface on TORAS~\cite{toras}. Details in Sec.~\ref{subsec_supp:annotation}.
    }
    \label{fig:annotation}
\end{figure*}

In this work, we present \dataset\ (WTC for short) as a large-scale video OSC dataset comprising fine-grained intra-object state-change segmentation. While the train split of WTC (\stat{$\sim$17k} clips) relies on the pseudo-labeling procedure detailed in Sec.~\ref{subsec:pseudo_label_gen} \& \ref{subsec:constraints} for training SPOC 
(\ref{subsec:model_training}), to ensure evaluation is rigorous and reliable, we collect manual annotations for \stat{1162} eval clips from experienced professional human annotators.

Here, we describe the annotation process and guidelines followed for spatial-OSC annotations.
For the 10 \task verbs in HowToChange~\cite{xue2023vidosc} (see Table~\ref{tab:htc_vost_big}), we first gather all the clips from the evaluation set of HowToChange, totalling \stat{2787} clips. Then, we manually filter the clips, removing clips with high motion blur, no state changes, ambiguous actionable and transformed states, and so on. This leads to a final evaluation set of \stat{1001} clips in WTC-HowTo. To curate the WTC-VOST evaluation set, we consider those verbs which overlap with WTC-HowTo (chop, peel). We gather clips from both train/val splits of VOST~\cite{tokmakov2023vost}, while adopting a similar framework to filter out ambiguous and blurry clips. This yields \stat{155} unique clips in WTC-VOST.
In total, our evaluation set comprises \stat{1162} clips across both subsets.

Next, with the help of \stat{7} expert human annotators, 
the entire annotation process is conducted on TORAS~\cite{toras} over the duration of a month, totaling 350 annotation hours. The annotators are first given a thorough outline of \task OSCs and ways of annotating them, followed by multiple quizzes and sample sets to test their understanding. A screenshot of the annotation user interface is shown in 
Fig.~\ref{fig:annotation} (i). 
An overview of the general guidelines provided to the annotators is shown in Fig.~\ref{fig:annotation} (ii).

Given a video clip and the corresponding OSC (e.g. mashing potato), the annotators first review the clip, identifying the object in different states of change. Then they proceed to annotate the video frame-by-frame, marking actionable and transformed regions of the relevant object. Frames where this distinction is unclear are marked as ignored frames.  Ignored frames are not considered while computing the IoU metric during evaluation.
Specific guidelines for each verb are listed in Table~\ref{tab:ann_guidelines}.
Once the annotations are complete, they are further vetted by us, and any edits are sent back to the annotators for corrections.

The aforementioned annotation procedure provides us with a high-quality evaluation set for \dataset, enabling us to conduct robust evaluation for the \task OSC task.  This data will be released publicly to allow further progress on the new task.

\begin{table}[htbp]
\tablestyle{4pt}{1.1}
\centering
\begin{tabular}{cp{6.7cm}}
    \toprule
    \textbf{Verb} & \textbf{Guidelines} \\
    \midrule
    Chopping & 
        \textit{Actionable}: Whole fruits and large pieces
        \newline \textit{Transformed}: Only small pieces, slices, or rings
        \newline Note: 1) If in a video, an object is cut in half and further chopped into smaller pieces, only the smaller pieces are transformed. The big halves initially are marked as actionable. 2) Hands or tools like knife, etc are excluded from the segmentation\\
    
    Slicing & 
        Same as above\\
    
    Mincing & 
        In addition to the above, mincing specifically means chopping into tiny pieces, not slices. So intermediate steps like slices are marked actionable \\
    
    Grating & 
        \textit{Actionable}: Whole or chunked piece
        \newline \textit{Transformed}: Grated pieces
        \newline Note: Grater and other tools are ignored \\
    
    Shredding 
        & Same as above \\
    
    Mashing & 
        \textit{Actionable}: Whole fruit or large chunks
        \newline \textit{Transformed}: Mashed regions of fruit
        \newline Note: The visual distinctions can be subtle, so textural difference are used as guidance. E.g. When mashing potatoes, unmashed regions are chunky, mashed regions are smooth \\
    
    Crushing & 
        Same as above \\
    
    Peeling & 
        \textit{Actionable}: Unpeeled region
        \newline \textit{Transformed}: Peeled region
        \newline Note: Peel that is detached from fruit is not segmented under either category. E.g. banana peel after being detached from fruit is not actionable or transformed, it is excluded from the segmentation \\
    
    Melting & 
        \textit{Actionable}: Unmelted solid region
        \newline \textit{Transformed}: Melted liquid or paste-like region \\
    
    Coating & 
        \textit{Actionable}: Object region plain and uncoated by external substance
        \newline \textit{Transformed}: Region coated with substance – e.g. bread coated with jam, nutella, butter, etc
        \newline Note: a) Tools used for coating like spoon, etc are excluded as much as possible unless it is present on the object and prominently coated.
        b) Only the object of interest is segmented as transformed if coated, not the coating substance if it is stand-alone. E.g. containers containing the substance are not segmented as transformed.
        \\
    \bottomrule
\end{tabular}
\caption{\textbf{Verb-specific annotation guidelines for \dataset. } General guidelines are in Fig.~\ref{fig:annotation} (i).
} 
\label{tab:ann_guidelines}
\end{table}

\subsection{Dataset Analysis}
\label{subsec_supp:dataset_analysis}

\begin{figure*}[htpb]
    \centering
    \includegraphics[width=\linewidth]{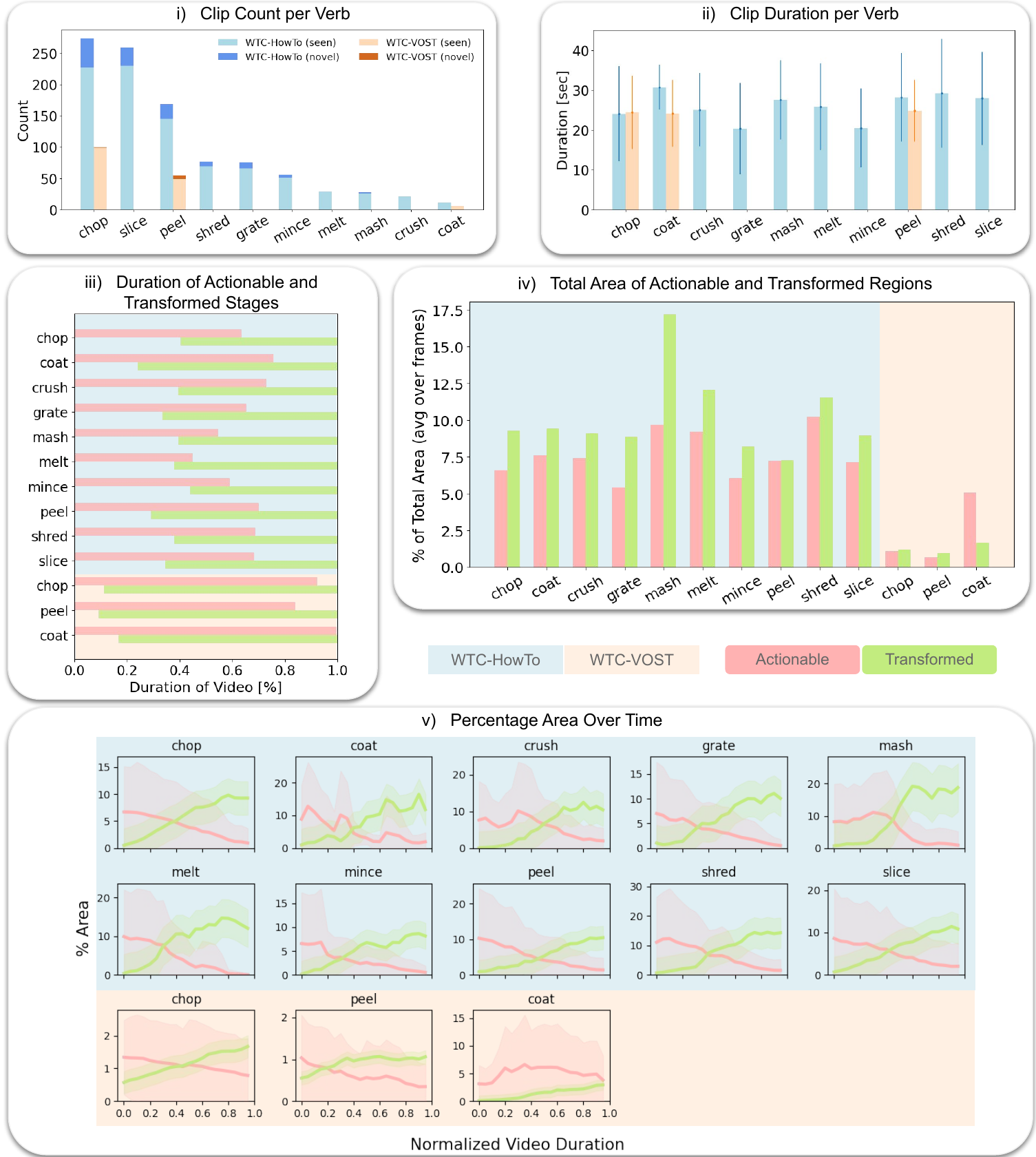}
    \caption{\textbf{Analysis of \dataset\ dataset along different axes. }
    \textbf{i) Clip counts per verb:} distribution of seen-novel object counts in WTC-HowTo and WTC-Vost. Object taxonomoy in Table~\ref{tab:osc_taxonomy}.
    \textbf{ii) Clip duration per verb:} clips are 20 seconds long on average across all verbs.
    \textbf{iii) Duration of actionable and transformed phases:} some verbs see a large overlap between actionable and transformed stages (e.g. coat, peel), indicating a slower and longer transition period. Others see a smaller overlap (e.g. melt), indicating a quicker transition from actionable to transformed regions.
    \textbf{iv) Area of actionable and transformed regions:} transformed regions occupy a larger area compared to actionable regions, likely indicating a larger surface area due to disintegration of the original object (e.g. chop, grate, mash). WTC-VOST has overall smaller areas compared to WTC-HowTo since the head-mounted camera captures a more zoomed out view compared to close-up shots in the latter.
    \textbf{v) Progression of actionable and transformed regions over time:} with time, the area of actionable regions decrease, while areas of transformed regions increase, highlighting the natural progression of OSC dynamics present in our annotations.
    }
    \label{fig:dataset_analysis}
\end{figure*}

We conduct a thorough analysis of our proposed \dataset dataset along different axes, highlighting its wide-ranging and diverse nature. A summary of the analysis is in Fig.~\ref{fig:dataset_analysis}. We explain each of the properties in detail below.

\custompar{Clip counts per verb}
We show the distribution of seen-novel object counts in WTC-HowTo (10 verbs) and WTC-Vost (3 verbs) in Fig.~\ref{fig:dataset_analysis} (i). The OSC object taxonomy is in Table~\ref{tab:osc_taxonomy}. Novel objects are not present for 3 verbs in WTC-HowTo (coat, crush, melt), hence only seen objects are reported. We see that our dataset comprises a wide collection of objects across both subsets.
Our dataset allows for testing models for generalization across diverse object classes (e.g. while apple is a commonly occurring fruit in HowTo chopping videos, date is relatively low-frequency).

\custompar{Clip duration per verb} 
We show the duration of clips for every verb in Fig.~\ref{fig:dataset_analysis} (ii).
Clips are 25 seconds long on average across all verbs with a standard deviation of roughly 10 seconds. Our minimum clip duration is \stat{3} seconds, while the maximum is \stat{96} seconds long.

\custompar{Duration of actionable and transformed phases} 
Fig.~\ref{fig:dataset_analysis} (iii) shows the duration of actionable and transformed stages for each verb. The actionable/transformed stage is those time-steps where an actionable/transformed mask is present in the annotation. We notice that some verbs have a large overlap between actionable and transformed stages (e.g. coat, peel), indicating a slower and longer transition period. Others see a smaller overlap (e.g. melt), indicating a quicker transition from actionable to transformed regions. This finding supports our intuitive understanding of the nature of these OSCs. Activities like chopping and grating are more drawn-out, involving human effort to act on the object and change it in slower stages. On the other hand, activities like melting ghee or butter can proceed very quickly on account of automatic catalysts like heat, requiring little human time and effort.

\custompar{Area of actionable and transformed regions} 
Fig.~\ref{fig:dataset_analysis} (iv) shows the area of actionable and transformed regions for each verb. We see that transformed regions occupy a larger area compared to actionable regions, likely indicating a larger surface area due to disintegration of the original object. For instance, chopping, slicing and grating disintegrate a whole fruit into multiple smaller pieces, considerably increasing surface area. This change is stark for mashing, where the volumetric object region is transformed a into flattened surface area. When comparing WTC-HowTo and WTC-VOST, the latter has overall smaller areas owing to the head-mounted camera, which captures a more zoomed out view compared to close-up shots in the former.

\custompar{Progression of actionable and transformed regions over time} 
Another interesting aspect of our dataset is that due to the spatial segmentation annotation, we can directly track the behavior of the actionable and transformed regions over time. Unlike frame-level labels in ~\cite{souvcek2022look,xue2023vidosc} or state-agnostic segmentations in ~\cite{tokmakov2023vost,Yu_2023_vscos}, our \task OSC task has the unique benefit of fine-grained \textit{spatial} and \textit{temporal} state-change maps. Making use of this, we track the area of actionable and transformed regions over time and present averaged results across all clips per verb in Fig.~\ref{fig:dataset_analysis} (v). 

We observe that with time, the area of actionable regions decrease, approaching 0, while those of transformed regions increase, highlighting the natural progression of OSC dynamics present in our annotations. We also show the standard deviation in a lighter shade. Notice how the standard deviation for transformed regions tapers down to lower values with time, with the reverse being true for actionable. When comparing WTC-HowTo and WTC-VOST, we observe that the former often sees activities reaching completion (transformed area close to 0), while the latter sees higher values. This is because WTC-VOST comprises continuously captured real-time videos, where the activity seldom reaches completion within the short clip duration. However, the natural progression of  shrinking actionable regions and expanding transformed regions is uniformly observed across both subsets and across all verbs.

\subsection{Dataset Samples} 
\label{subsec_supp:dataset_samples}

\begin{figure*}[htpb]
    \centering
    \includegraphics[width=\linewidth]{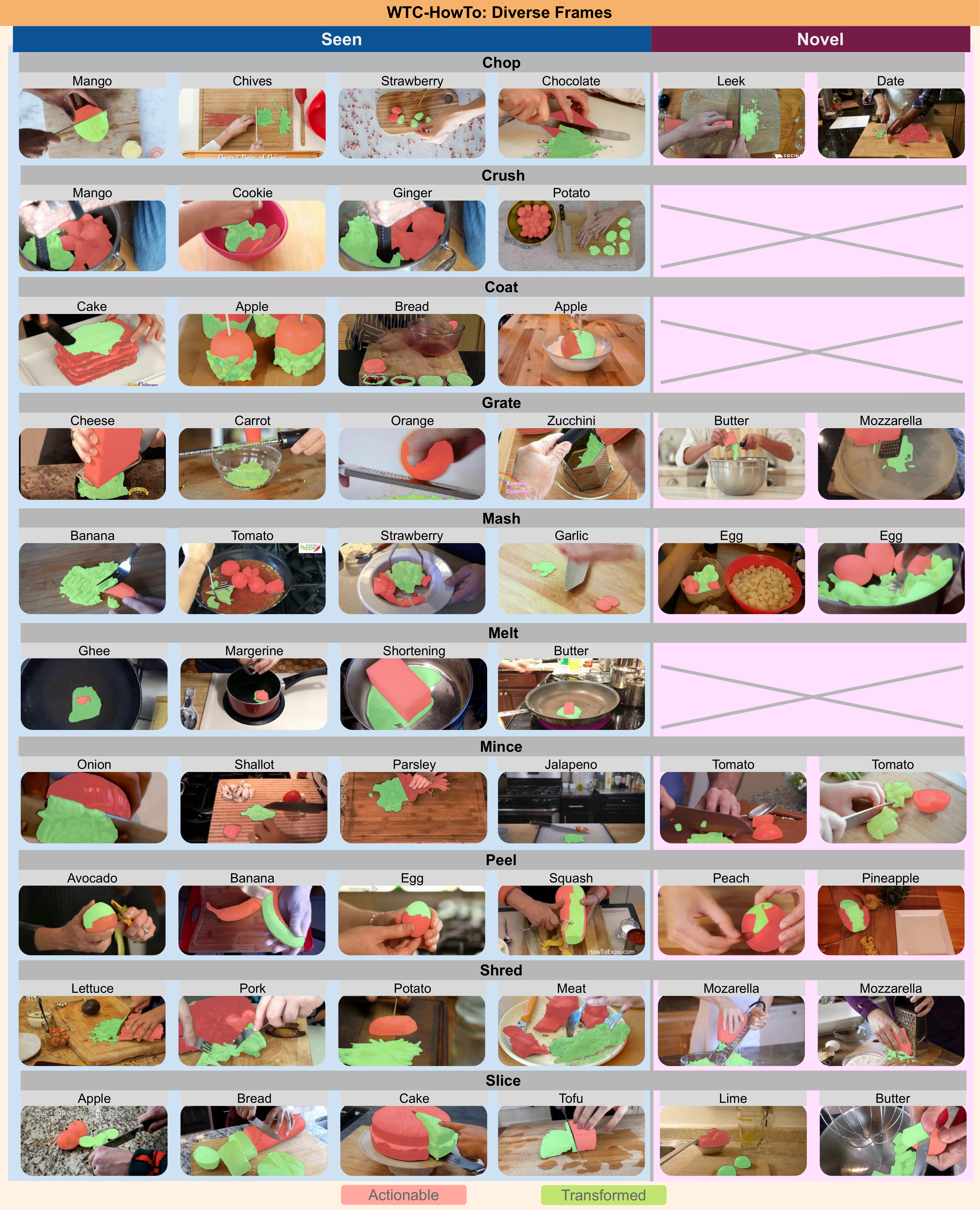}
    \caption{\textbf{Annotation samples from \dataset-HowTo. }
    Diverse frame samples from the HowToChange~\cite{xue2023vidosc} subset of our proposed \dataset\ dataset. We show samples from both seen and novel object splits across all verbs and distinct nouns. Note that for three verbs (crush, coat, melt), we only have seen objects in the evaluation set, and hence omit novel object samples.
    }
    \label{fig:dataviz_htc}
\end{figure*}

\begin{figure*}[htpb]
    \centering
    \includegraphics[width=\linewidth]{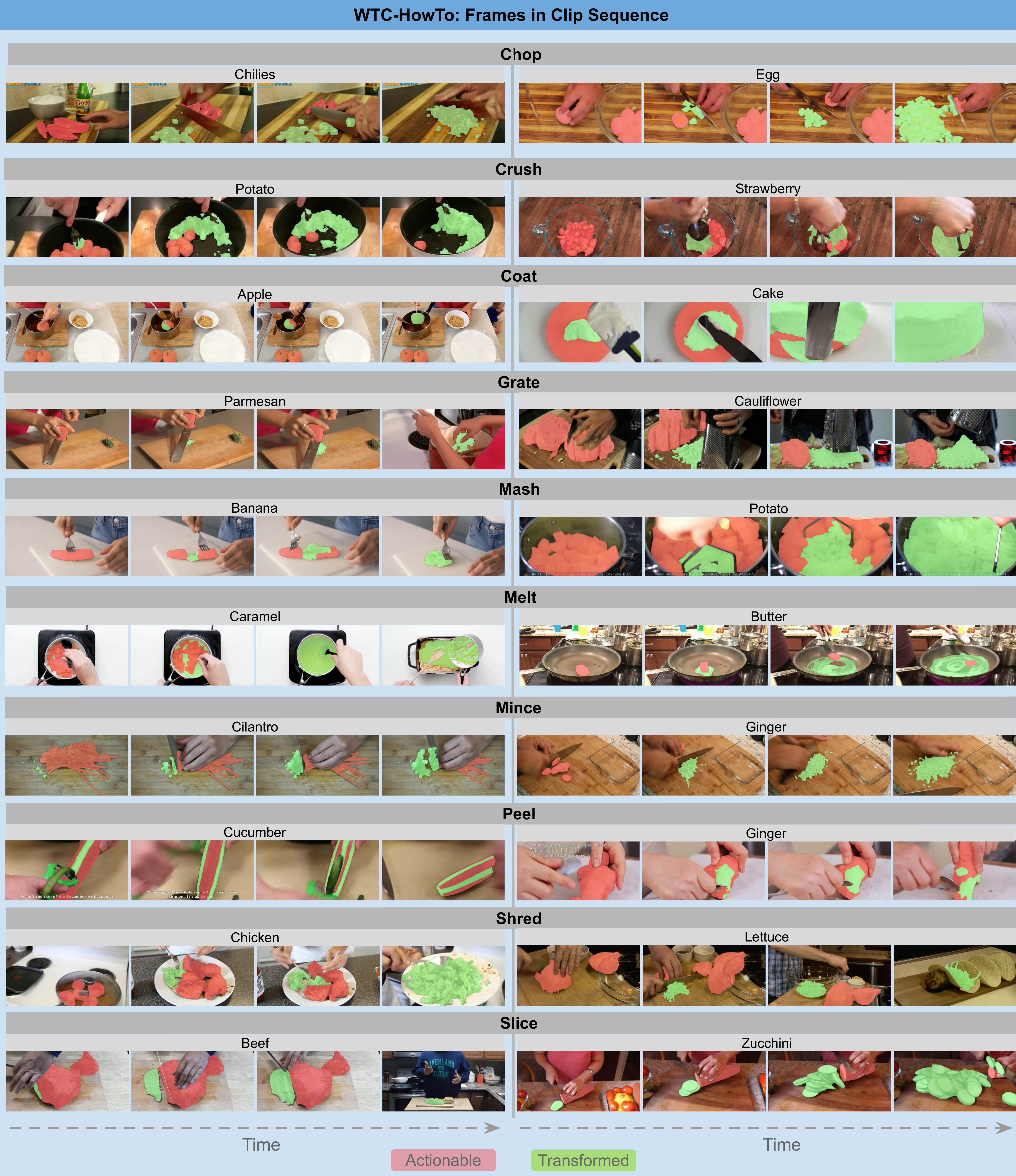}
    \caption{\textbf{Annotation clip sequences from \dataset-HowTo. }
    We show sample frames from single clip sequences from the HowToChange~\cite{xue2023vidosc} subset of our proposed \dataset\ dataset. Notice that with the passage of time, actionable regions progressively change into transformed regions, following the natural progression of the OSC.
    }
    \label{fig:dataviz_htc_clips}
\end{figure*}

\begin{figure*}[htpb]
    \centering
    \includegraphics[width=\linewidth]{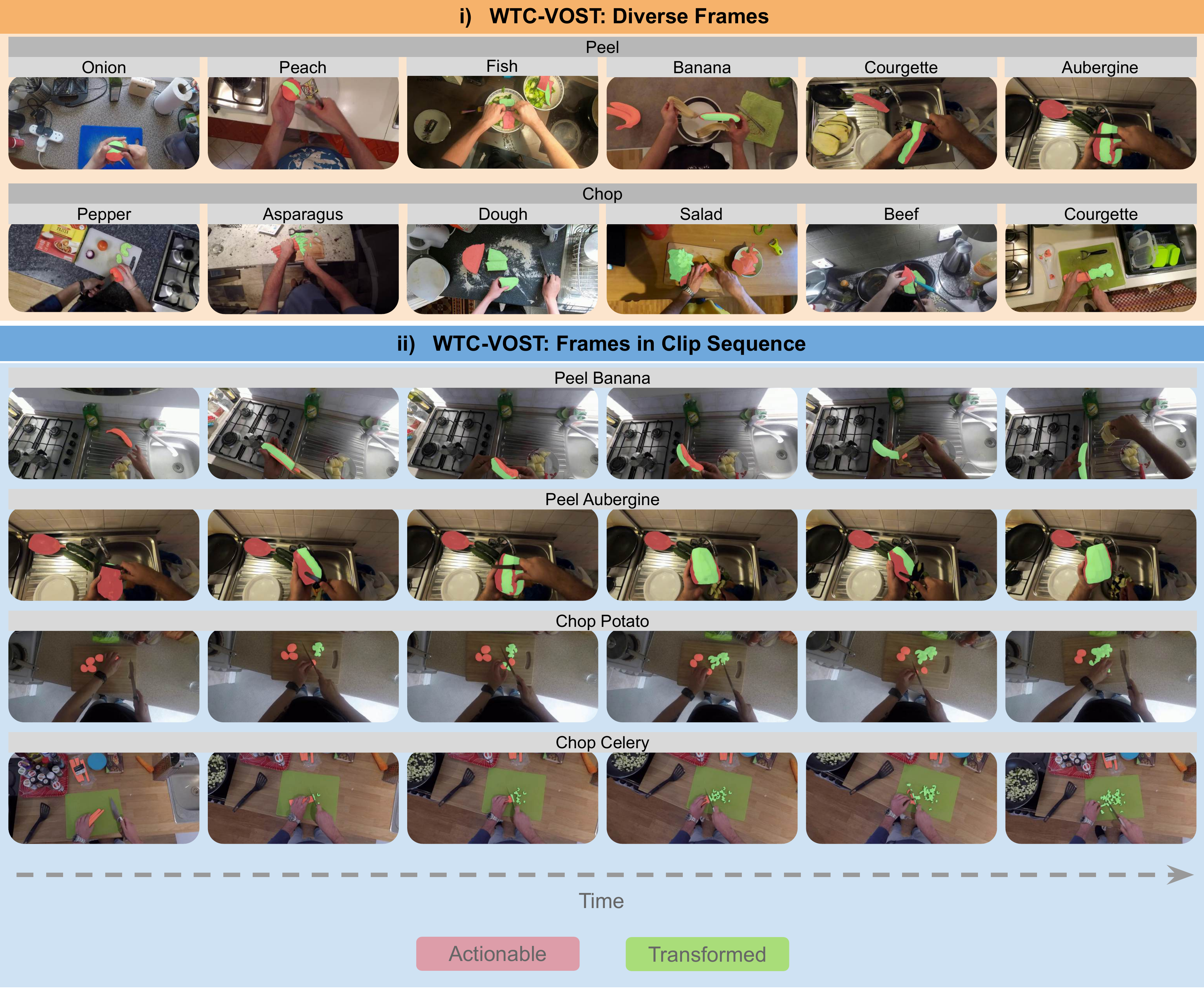}
    \caption{\textbf{Annotation samples and clip sequences from \dataset-VOST. }
    We show sample frames \textbf{(i)} and clip sequences \textbf{(ii)} from the VOST~\cite{tokmakov2023vost} subset of our proposed \dataset\ dataset. This is a more challenging out-of-distribution dataset comprising continuously captured egocentric videos of human activities.
    }
    \label{fig:dataviz_vost}
\end{figure*}

We present representative samples from our WhereToChange evaluation set across all subsets (WTC-HowTo/WTC-VOST), data splits (Seen/Novel) and verbs in Figs.~\ref{fig:dataviz_htc},~\ref{fig:dataviz_htc_clips},~\ref{fig:dataviz_vost}. Fig.~\ref{fig:dataviz_htc} shows a diverse collection of annotation samples for the 10 verbs containing spatial-OSCs in WTC-HowTo. Notice the fine-grained nature of the annotations, encapsulating precise boundaries of actionable and transformed regions (e.g. chopping chives, grating carrot, mashing tomato, and so on).
In Fig.~\ref{fig:dataviz_htc_clips}, we depict frames within a clip sequence. We observe that with the passage of time, actionable regions progressively change into transformed regions, following the natural progression of the OSC.
In Fig.~\ref{fig:dataviz_vost}, we show frames and sequences from WTC-VOST, the out-of-distribution subset of \dataset. In this more challenging dataset comprising continuously captured egocentric videos, we yet again observe the natural OSC dynamics observed earlier viz smooth progression of actionable to transformed regions. 

In summary, \dataset, is a wide-ranging dataset comprising \task OSCs from a plethora of cooking activities. With fine-grained intra-object segmentations over a diverse set of seen and novel objects, we push the frontiers of video OSC understanding.

\section{Experimental setup}
\label{sec_supp:expt_setup}
 In this section, we discuss in detail our experimental setup including implementation details, metrics and baselines.

\subsection{Implementation details}
\label{subsec_supp:implementation}

\custompar{Pseudo-labeling}
We list hyperparameter settings for each component of the pseudo-labeling stage of SPOC in Table~\ref{tab:hyperparams}. We perform pseudo-labeling at 5 fps. In particular, we find that despite using a tracker~\cite{yang2022deaot}, running the detector every at frequent intervals is necessary in order to re-identify dropped objects. Since objects morph rapidly with time, the tracker can often fail to track the objects meaningfully. Hence, we run the detector every 10 frames (2 seconds) in the video clip.
For SAM, a 32 × 32 point grid is employed, and non-maximum suppression (NMS) with a threshold of 0.9 is applied to eliminate redundant mask proposals. 
In preliminary experiments on the choice of the CLIP vision encoder, we didn't find a significant performance difference between ViT-B/16 \& ViT-B/32. We chose ViT-B/32
for faster pseudo-label generation at large scale. However, SPOC would naturally benefit from stronger representations; future use of more advanced backbones is a clear opportunity for further improvement.

\custompar{Text Prompt}
For generating the text prompts for object detection, we automatically generate language labels with LLMs. 
We prompt ChatGPT-4 to generate actionable and transformed 
phrases for various OSCs following a few manual examples (e.g. prompt: "OSC: chopping avocado, actionable: whole avocado, transformed: chopped avocado pieces. What is actionable and transformed label for OSC: slicing tomato?")

\custompar{Model Training}
During training, our video model consists of a 3-layer transformer encoder (512 hidden dimension, 4 attention heads) and a 1-layer MLP decoder.
While training the transformer model, we process 16 frames at a time sampled at 1 fps, w/ 16 global frame features and up to 4 mask features per frame, with both mask and time-positional embeddings.
While the pseudo-labels include an ``ambiguous'' category, the SPOC model classifies each proposal into one of three states---actionable, transformed, or background. The ambiguous label is handled by preventing loss backpropagation for such proposals, ensuring they do not influence model training.
We employ the AdamW optimizer with a learning rate of 1e-4 and a weight decay of 1e-4. Models are trained using a batch size of 64, over 50 epochs.
Our inference speed is ~4 FPS ( on an Nvidia RTX 6000 GPU), with most time spent on mask proposal generation via SAM.
Our model parameters are as follows: Trainable -- SPOC Transformer (9.7M); Non-trainable -- CLIP (88M), SAM (636M), DeAOT (49M); 
This is comparable to existing VOS models like DeAOT, and significantly faster than heavier models such as SAM (0.5 FPS, 636M params).

\begin{table}[htpb]
\tablestyle{6pt}{1.1}
\centering
\begin{tabular}{ll}
    \toprule
    \textbf{Hyperparam} & \textbf{Value}\\
    \midrule

    \rowcolor{LighterGray}
    \multicolumn{2}{c}{Grounding-Dino~\cite{liu2023gdino}} \\
    box threshold & 0.35 \\
    text threshold & 0.5 \\
    box size threshold & 0.7 \\

    \rowcolor{LighterGray}
    \multicolumn{2}{c}{SAM~\cite{ren2024gsam}} \\
    model type & vit\_h \\
    prompt type & bounding box \\

    \rowcolor{LighterGray}
    \multicolumn{2}{c}{Clip~\cite{clip}} \\
    model & ViT-B/32 \\

    \rowcolor{LighterGray}
    \multicolumn{2}{c}{DeAOT~\cite{yang2022deaot}} \\
    phase & PRE\_YTB\_DAV \\
    model & r50\_deaotl \\
    long term mem gap & 9999 \\
    max len long term & 9999 \\

    \rowcolor{LighterGray}
    \multicolumn{2}{c}{SamTrack~\cite{cheng2023samtrack}} \\
    sam gap & 10 \\
    min area & 50 \\
    max obj num & 255 \\
    min new obj iou & 0.8 \\

    \bottomrule
\end{tabular}
\caption{\textbf{Hyperparameter settings for different components during pseudo-labeling. } Details in Sec.~\ref{sec_supp:expt_setup}.
} 
\label{tab:hyperparams}
\end{table}

\subsection{Baselines}
\label{subsec_supp:baselines}

We evaluate several state-of-the-art segmentation baselines for the task of spatially progressing object state change segmentation. These baselines fall into two broad categories: pixel-based (MaskCLIP\cite{zhou2022maskclip}, MaskCLIP+\cite{zhou2022maskclip}) and object-centric (GroundedSAM\cite{ren2024gsam}, SAMTrack\cite{cheng2023samtrack}, DEVA~\cite{cheng2023deva}).
\begin{itemize}
    \item \textbf{Pixel-based} methods perform dense segmentation, classifying each pixel in the image into one of three regions: actionable, transformed, or background.
    \item \textbf{Object-centric} methods first employ off-the-shelf open-vocabulary detectors to identify relevant object regions before classifying them into the same three categories.
\end{itemize}
This structured evaluation allows us to compare the effectiveness of different approaches in capturing fine-grained state changes within objects.

We use official implementations of all baseline methods with their default hyperparameter setting values. Below, we explain each baseline in detail.

\custompar{MaskCLIP~\cite{zhou2022maskclip}}
This is a state-of-the-art semantic segmentation method that leverages CLIP for pixel-level dense prediction to achieve annotation-free segmentation. 
To this end, keeping the pretrained CLIP weights frozen, they make minimal adaptations to generate pixel-level classification. Furthermore, MaskCLIP+ uses the output of MaskCLIP as pseudo-labels and trains a more advanced segmentation network. In the main paper Table~\ref{tab:htc_vost_big}, 
we report zero-shot results using MaskCLIP ViT-B/16. For the foreground prompt, given an OSC, e.g. 'chopping avocado', we prompt with 'chopped avocado pieces' and 'whole avocado'. In addition, following the original work, we also include background classes that are commonly present in the scene (e.g. person, hand, table, knife, etc) since it greatly improves the performance of MaskCLIP.

\custompar{MaskCLIP+~\cite{zhou2022maskclip}}
Further, we train MaskCLIP+, using an advanced pixel-level segmentation model (Deeplabv2-ResNet101) on the pseudo-labels generated by MaskCLIP on WTC-HowTo to obtain trained metrics. While training generally improves performance, we note that the rest of the object-centric methods (that use language-prompted detection followed by classification) still fare better compared to MaskCLIP+. This is because MaskCLIP still contains much noise in non-object regions owing to dense pixel-level predictions, while object-centric model predictions are mostly contained within relevant objects.

\custompar{Grounded-SAM~\cite{ren2024gsam}}
We first use GroundingDino~\cite{liu2023gdino} to obtain text-prompted object bounding boxes. For e.g. for the OSC ``chopping avocado'', we use the prompts ``whole avocado'' and ''chopped avocado pieces`` to prompt GroundingDino to obtain actionable and transformed boxes. These boxes are then used to prompt SAM~\cite{kirillov2023sam} to obtain masks. For the trained version, we use the pseudolabels generated by GroundedSAM to train our transformer model (Sec. 3.4). We find that the model fails to learn any reasonable information about states due to the poor pseudo-label quality that closely resembles random state assignment (compare with Random Label in Table~\ref{tab:htc_vost_big}).

\custompar{SamTrack~\cite{cheng2023samtrack}}
In addition to intermittently detecting relevant objects, this method also uses DeAOT~\cite{yang2022deaot}, a tracker that tracks segmented object regions for the remainder of the video. We notice that large structural changes in the object can affect tracking, hence it becomes important to run the detector routinely to re-detect the dropped object and pass it back to the tracker.

\custompar{Random Label}
In this baseline, we consider the object masks generated by SamTrack, however we assign ``actionable'' and ``transformed'' labels randomly throughout the video. A lot of the pseudo-labels generated by the baseline methods (e.g. GroundedSAM, SAMTrack, DEVA) all perform similarly to random label assignment. This underscores the challenging nature of our spatially-progressing OSC task---distinguishing between object states is a limiting factor of existing detection and segmentation methods.

\custompar{DEVA~\cite{cheng2023deva}}
This is a decoupled video segmentation approach that uses XMem~\cite{cheng2022xmem} to store and propagate masks. Similar to the pipeline in SAMTrack, an object detection~\cite{liu2023gdino} and segmentation~\cite{kirillov2023sam} pipeline is adopted to first detect relevant object masks. These masks are later aggragated and propagated through the rest of the video using XMem. Akin to the above, we run routine detection (every 10 frames) to recapture dropped objects.
The default setting in DEVA~\cite{cheng2023deva} generates a large number of mask proposals, drastically increasing run-time and GPU memory. For clips where we encounter memory overflows, we reduce max number of tracked objects from 100 to 20. We keep rest of the hyperparameters the same.

\begin{figure*}[t]
    \centering
    \includegraphics[width=0.8\linewidth]{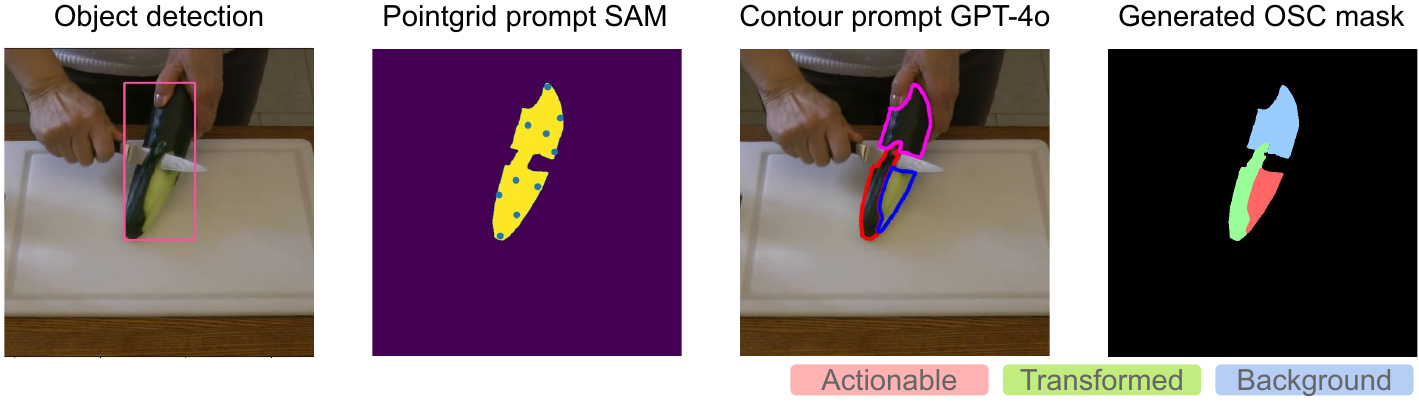}
    \caption{\textbf{GPT-4o labeling pipeline.} 
    We use \cite{ren2024gsam} for object segmentation, extract intra-object parts via a point-grid prompt, and query GPT-4o to classify each part into one of actionable, transformed, or background. We notice two issues that arise: 1) parts are not reliably captured, often being hallucinated where none exist, or clubbing multiple distinct regions together 2) GPT-4o labels are less reliable, consistently flipping between the three labels throughout the video for the same segment.}
    \label{fig:gpt-4o_pipeline}
\end{figure*}

\custompar{GroundedSAM~\cite{ren2024gsam} + GPT-4o~\cite{openai2023gpt4}}
This part-centric baseline combines GroundedSAM for object segmentation with GPT-4o for intra-object classification. We sample a grid of points within each detected object mask and point-prompt SAM to generate candidate parts. These regions are then contoured and passed to GPT-4o with task-specific prompts to label each as \{actionable, transformed, background\}.
For example, for peeling cucumber: “What does the contoured region correspond to? (a) unpeeled cucumber region, (b) peeled cucumber region, (c) not a cucumber).” See Fig.~\ref{fig:gpt-4o_pipeline}.  
To maintain prompt reliability, we limit each image query to at most four contours, issuing multiple prompts if an image contains more parts. This baseline evaluates whether a state-of-the-art VLM can assign fine-grained, state-sensitive labels to object regions, beyond whole-object segmentation.

We find two main challenges on in-the-wild WTC videos:
(a) Intra-object proposals are unstable—SAM2 often over-segments subtle texture changes or deformations, hallucinating “parts” where none exist. This is especially harmful for verbs like cutting, where the model may infer a slice despite no slicing occurring.
(b) Labeling consistency over time is weak—the same region frequently flips between actionable and transformed, showing poor progress modeling. By contrast, SPOC’s causal ordering and temporal aggregation substantially reduce such errors.

We notice that object-centric methods outperform pixel-based methods in our task. They have the advantage of well-defined object regions provided by the object detection~\cite{liu2023gdino} and segmentation~\cite{kirillov2023sam} pipeline, whereas pixel-based methods often have a greater degree of noise in the predictions. However, object-centric methods also suffer the disadvantage of being unable to assign two separate labels to regions within one detection. While the current SPOC model also follows an object-centric classification paradigm during training, the pseudo-labels could also be used to train dense pixel-based segmentation models. Future work could integrate object-centric and pixel-based techniques for optimal results.

\subsection{Metrics} 
\label{subsec_supp:metrics}

Following prior segmentation works~\cite{Perazzi2016davis,tokmakov2023vost,Yu_2023_vscos,ren2024gsam}, we report mean IoU scores as a measure of segmentation performance. The reported mIoU is the average over actionable and transformed regions:
\begin{equation}
    \label{iou}
    mIoU = \frac{mIoU_{act} + mIoU_{trf}}{2}
\end{equation}
Following prior segmentation works~\cite{tokmakov2023vost,shen2024rosa} that evaluate on video OSC datasets, we do not report boundary f-measure. This is because the objects undergoing OSC often change dramatically with constantly evolving boundaries, making it challenging to track and ascertain object boundaries.

\section{SPOC Ablations}
\label{sec_supp:ablations}

\begin{figure}[htpb]
    \centering
    \includegraphics[width=\linewidth]{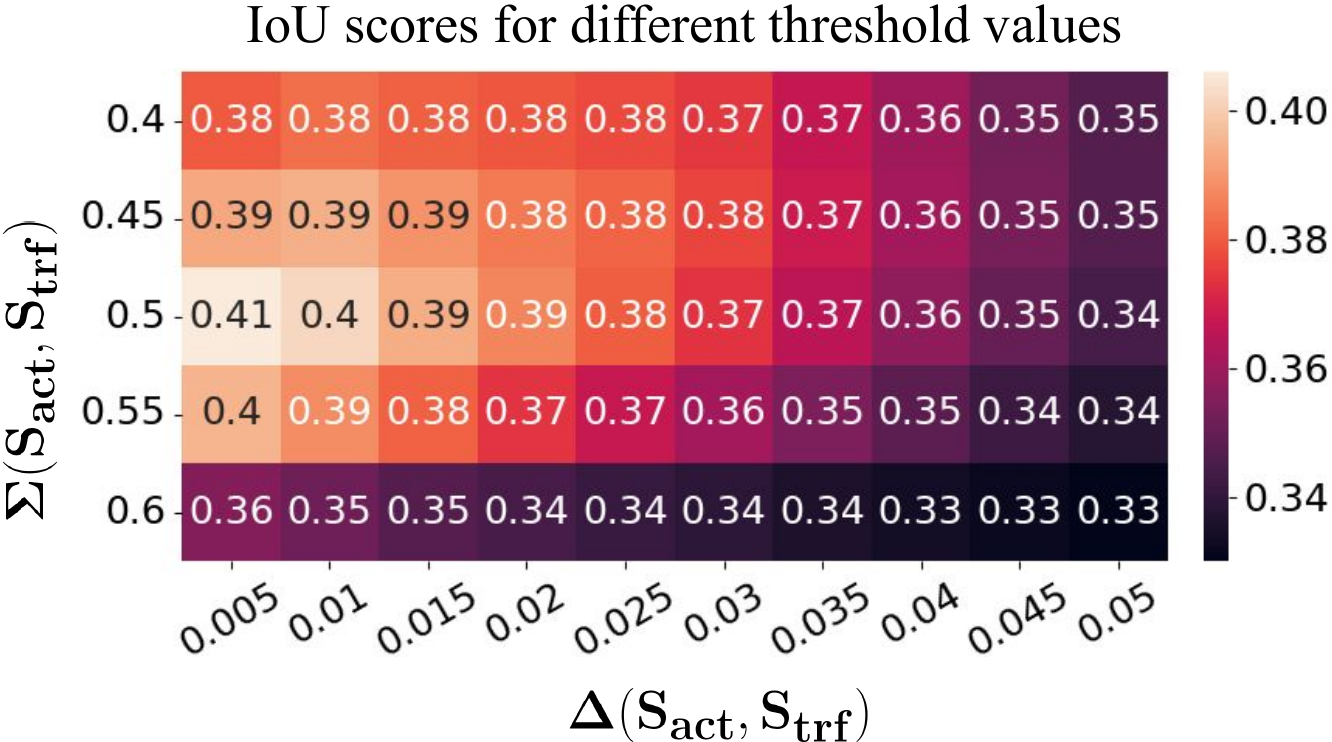}
    \caption{\textbf{IoU scores for different CLIP similarity threshold values. }
    We plot a heatmap of $\sum(S_{act},S_{trf})$ and $\Delta(S_{act},S_{trf})$, CLIP similarity threshold values in Eq.~\ref{eq:clip_thr}. This heatmap shows aggregate values across all verbs. For our pseudo-labels, we choose best threshold values computed for each verb.
    }
    \label{fig:clip_thr_heatmap}
\end{figure}

\custompar{Effect of varying CLIP thresholds}
In Sec.~\ref{subsec:pseudo_label_gen}, we make use of CLIP's~\cite{clip} vision-language embeddings to pseudo-label each mask proposal into one of ($s_{act}$, $s_{trf}$, $s_{amb}$, $s_{bg}$) based on similarity threshold values between the mask-vision and OSC-text embeddings. Eq.~\ref{eq:clip_thr} follows a thresholding mechanism using $\Sigma{s_{act},s_{trf}}$ and $\Delta{s_{act},s_{trf}}$ values to set the pseudo-labels. We run a grid-search over $\Sigma$ and $\Delta$ for each verb and compute IoU metrics on the generated pseudo-labels to choose the best threshold values.
Fig.~\ref{fig:clip_thr_heatmap} shows a heatmap of $\sum(S_{act},S_{trf})$ and $\Delta(S_{act},S_{trf})$ and their respective IoU values. This heatmap shows aggregate values across all verbs. For our pseudo-labels, we choose best threshold values computed for each verb. This aggregates to a $\Sigma$ value of 0.5 and a $\Delta$ value of 0.01.

\custompar{Importance of regular detection}
We use an object detector and tracker to detect and track the relevant object for the duration of the state-change video. However, since the object often undergoes dramatic transformations such as shape, color or texture changes, the tracker might struggle to reliably track the object for prolonged periods. 
Further, while our task is a form of VOS, our formulation is designed to handle real-world instructional videos—where jump cuts are common—and the object may move in and out of view.
Rather than artificially require continuity (which would severely limit data diversity),
we mitigate both the above issues by running the detector every 10 frames, passing potentially dropped objects back to the tracker.
To ascertain the importance of regular detection, we run an ablation where we only detect the relevant object once at the start of the video while relying on the tracker's predictions for the remainder of the frames. 
As seen in Table~\ref{tab:detection}, only first-frame detection and tracking yields 0.372 mIoU on WTC-HowTo---10\% below SPOC (PL).
Hence, regular detection is necessary to maintain object identity during tracking.

\begin{table*}[h]
\tablestyle{4pt}{1.15}
\centering
\aboverulesep=0ex
\belowrulesep=0ex
\begin{NiceTabular}{l|c|cccccccccc}
    \toprule
    \textbf{Detection} & \multicolumn{11}{c}{\textbf{WTC-HowTo}} \\
     & \textbf{\underline{Mean}} & \textbf{Chop} & \textbf{Crush} & \textbf{Coat} & \textbf{Grate} & \textbf{Mash} & \textbf{Melt} & \textbf{Mince} & \textbf{Peel} & \textbf{Shred} & \textbf{Slice} \\
    \midrule
    only 1st-frame detection & 0.372 & 0.405 & 0.332 & 0.184 & 0.389 & 0.488 & 0.344 & 0.450 & 0.348 & 0.385 & 0.396  \\
    \rowcolor{LighterGray}
    every 10th frame detection & 0.411 & 0.438 & 0.351 & 0.237 & 0.429 & 0.527 & 0.416 & 0.493 & 0.373 & 0.418 & 0.428 \\
    \bottomrule
\end{NiceTabular}
\caption{\textbf{Importance of regular detection while tracking.}
To ascertain the importance of regular detection, we run an ablation where we only detect the relevant object once at the start of the video while relying on the tracker's predictions for the remainder of the frames. As seen, this degrades performance since the objects can often dramatically change shape, color, or texture. Hence regular detection is necessary to maintain object identity during tracking.
}
\label{tab:detection}
\end{table*}

\custompar{Importance of global-local representations}
As detailed in Sec.~\ref{subsec:model_training}, SPOC is trained using a combination of global frame-level Dino-v2~\cite{oquab2023dinov2} features and local mask-level CLIP~\cite{clip} features as inputs. To ablate the importance of each of these features, we train two variants of SPOC (global-only and local-only) which only use the frame-level features and mask-level features respectively. In the global-only setup, we pass bounding box locations as local inputs to differentiate between masks within the frame instead of CLIP features. We report results across 3 verbs in Table~\ref{tab:global_local}. As seen, the combination of global and local features yields the best performance, highlighting the importance of both granularities for our task.

\begin{table}[htpb]
\tablestyle{3.4pt}{1.1}
\centering
\begin{tabular}{lccc}
    \toprule
    \textbf{Model} & \textbf{Chop}  &  \textbf{Grate} &  \textbf{Peel} \\
    \midrule
    SPOC (global) & 0.423 & 0.469 & 0.401 \\
    SPOC (local) & 0.504 & 0.508 & 0.432 \\
    \rowcolor{LighterGray}
    SPOC (global+local) & \textbf{0.523} & \textbf{0.528} & \textbf{0.449} \\
    \bottomrule
\end{tabular}
\caption{\textbf{Importance of global and local features in SPOC model. } The combination of both frame-level global Dino features and mask-level local CLIP features yields the best performance.
} 
\label{tab:global_local}
\end{table}

\custompar{Upper-bound performance with oracle predictions}
We run an ablation to evaluate the upper-bound performance of the SPOC model with the existing object detector and mask tracking pipeline. Given mask proposals at each time-step, the trained SPOC model labels each into one of 3 categories: actionable, transformed or background. In our upper-bound experiment, we wish to gauge performance had SPOC made perfect predictions for each input proposal. To do this, we assign labels for each proposal from the ground truth annotations. If there is a large overlap of the proposal with $GT_{act}$, we assign $s_{act}$, for a large overlap with $GT_{trf}$, we assign $s_{trf}$, and $s_{bg}$ if there is no significant overlap with either. This serves as an oracle upper-bound for SPOC given no changes in detection and tracking. 

We report the mean IoU scores in Table~\ref{tab:upper_bound}. The upper-bound oracle reaches an mIoU of 0.65, as against the SPOC trained model at 0.502. This indicates that while SPOC (trained) improves over pseudo-labels (0.455), further improvements in the training architecture could enhance SPOC performance even further. However, larger improvements in the off-the-shelf detector and tracker are necessary to push beyond the upper-bound score of 0.65 mIoU.

\begin{table*}[h]
\tablestyle{4pt}{1.15}
\centering
\aboverulesep=0ex
\belowrulesep=0ex
\begin{NiceTabular}{l|c|cccccccccc}
    \toprule
    \textbf{Model} & \multicolumn{11}{c}{\textbf{WTC-HowTo}} \\
     & \textbf{\underline{Mean}} & \textbf{Chop} & \textbf{Crush} & \textbf{Coat} & \textbf{Grate} & \textbf{Mash} & \textbf{Melt} & \textbf{Mince} & \textbf{Peel} & \textbf{Shred} & \textbf{Slice} \\
    \midrule
    SPOC\ (pesudo-labels) & 0.455 & 0.461 & 0.383 & 0.447 & 0.447 & 0.566 & 0.418 & 0.500 & 0.417 & 0.453 & 0.458  \\
    \rowcolor{LighterGray}
    SPOC\ (trained) & 0.502 & 0.523 & 0.422 & 0.526 & 0.528 & 0.610 & 0.425 & 0.541 & 0.449 & 0.503 & 0.494 \\
    \rowcolor{LighterGray}
    SPOC\ (oracle) & \textbf{0.650} & \textbf{0.696} & \textbf{0.565} & \textbf{0.633} & \textbf{0.596} & \textbf{0.742} & \textbf{0.623} & \textbf{0.708} & \textbf{0.615} & \textbf{0.642} & \textbf{0.680} \\
    \bottomrule
\end{NiceTabular}
\caption{\textbf{Upper-bound oracle performance on WhereToChange-HowTo.}
We assign labels to input mask proposals by comparing with ground truth annotation masks to obtain upper-bound oracle performance for SPOC. 
While the trained model improves over the pseudo-labels (including constraints), improvements in the training architecture could enhance SPOC performance even further. However, larger improvements in the off-the-shelf detector and tracker are necessary to push beyond the upper-bound score of 0.65 mIoU. Details in Sec.~\ref{sec_supp:expt_setup}.
}
\label{tab:upper_bound}
\end{table*}

\begin{figure*}[t]
    \centering
    \includegraphics[width=\linewidth]{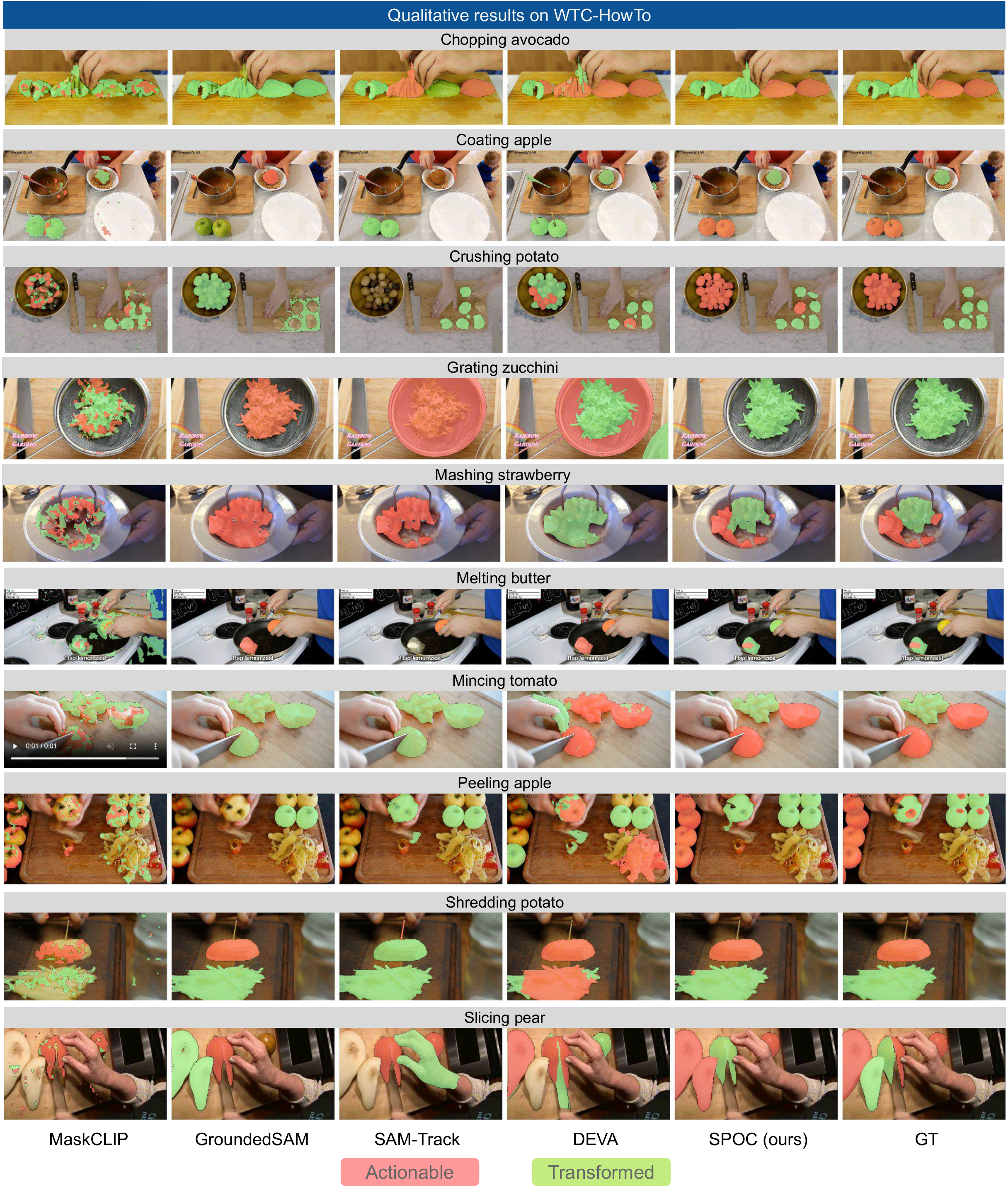}
    \caption{\textbf{Qualitative results on WTC-HowTo subset. } 
    In addition to the results in Fig.~\ref{fig:main_results}, we provide additional qualitative results comparing SPOC with all baselines across all verbs in WTC-HowTo. We observe that SPOC predictions are more sensitive to object states, while the baselines are more prone to mis-identifying actionable and transformed object regions.
    }
    \label{fig:results_htc_supp}
\end{figure*}

\custompar{Additional qualitative results}
In addition to the results in Fig.~\ref{fig:main_results}, we provide additional qualitative results comparing SPOC with all baselines across all verbs from WTC-HowTo in Fig.~\ref{fig:results_htc_supp}. We observe that SPOC predictions are more sensitive to object states, while the baselines are more prone to mis-identifying actionable and transformed object regions.

\section{Downstream Activity Progress}
\label{sec_supp:progress}
In Sec.~\ref{sec:expts}, we proposed activity progress tracking as a downstream task to gauge the utility of \task OSCs. We now provide details on the baselines and metrics.

\subsection{Progress Baselines}
Activity progress-monitoring is vital for AR/MR and robotics applications. In this regard, prior approaches~\cite{ma2022vip,ma2023liv} to robot learning have sought to learn goal-based representations that can be recast as reward functions for training robots. We use these state-of-the-art approaches as baselines for our progress-monitoring task.

\custompar{VIP}
Ma \etal~\cite{ma2022vip} introduced \textsc{VIP}, a self-supervised visual reward and representation learning technique trained on Ego4D~\cite{grauman2022ego4d} for downstream robot tasks. They key idea was to train an implicit value function that learns smooth and regular embeddings of egocentric video frames. From these frame-level embeddings, a goal-based reward function was formulated to be the goal-embedding distance difference at each time-step. 
Specifically, the reward at time-step $t$ is defined as:
\begin{equation}
    \label{eq:progress_vip}
    R(t) = \Phi(s_t) - \Phi(s_{g}) 
\end{equation}
where $\Phi(s_t)$ is the embedding at time-step $t$, and $\Phi(s_g)$ is the embedding at time-step $T$ i.e. goal-image $s_g$.
We refer the reader to ~\cite{ma2022vip} for further details. For our task, we supply the last image in the sequence as $s_g$. This yields goal-conditioned reward curves that double as progress curves.

\custompar{LIV}
As a follow up to VIP, Ma \etal~\cite{ma2023liv} introduced LIV as a dual language-image pretrained model that is capable of learning reward curves conditioned on both language and image embeddings. To adopt it for our task, we supply the language goal to be relevant OSC (e.g. chopping carrot), and the image goal is the last frame in the clip sequence as defined earlier.

\custompar{LIV-finetune}
Following the recommendations of the LIV paper, we finetune the pretrained LIV model on our WTC-HowTo training split and report results. We finetune for a total of 40 epochs and report results for the checkpoint yielding the best performance on the progress task.

\subsection{Progress Metrics}
We formulate progress metrics to evaluate two key properties that we would like to observe in the activity progress curves. First, since we consider irreversible state changes, the progress curves need to have a monotonic nature as the activity proceeds. Second, once the activity is complete, there should be no more changes in the curve values. This indicates the end of the task and should be denoted by stagnant curves. We now elaborate on each of the metrics.

\custompar{Kendall's Tau}
As stated above, activity progress curves need to reflect the monotonic nature of irreversible OSC progression. Here, we consider ideal curves to begin from a value of 1 (start of the activity, no progress) and finish at 0 (end of the activity, task completion). In other words, as the activity proceeds in time, the progress curve should smoothly proceed from 1 to 0, with a monotonically decreasing behavior for the duration of the activity. Kendall's Tau is a statistical measure that can determine how well-aligned two sequences are in time. It has been used in prior video alignment works~\cite{dwibedi2019tcc,Donahue_2024_CVPR} to measure temporal alignment in video frames.

For our task, we compute Kendall’s Tau ($\tau$) of the curve as a measure of monotonicity. Since it evaluates the ordinal association between two quantities, it can be adapted to measure monotonicity in a single clip sequence as follows:
\begin{equation}
    \label{eq:progress_tau}
    \small{
        \tau = \frac{\text{number of increasing pairs} - \text{number of non-increasing pairs}}{\text{all possible pairs}}
        }
\end{equation}
where a datapoint in the curve is the progress value at a particular time-step in the sequence. A pair-wise metric compares pairs of time-steps throughout the sequence.
In the equation above, a $\tau$ value of +1 indicates a perfectly monotonically increasing sequence, -1 indicates a perfectly monotonically decreasing sequence, and 0 indicates no monotonicity. Hence, in our case, a high negative $\tau$ value is an indicator of good task progress. 

In Table~\ref{tab:progress} of the main paper, OSC-based curves (SPOC-annotations and SPOC-model) have lower $\tau$ values compared to VIP~\cite{ma2022vip} and LIV~\cite{ma2023liv}, the goal-based representation learning baselines.
Contrary to the findings in the LIV paper regarding improved results with finetuning, we observe no significant difference in performance with or without finetuning. This underscores the challenging nature of our \task OSC task. Approaches which merely rely on goal-images while placing no special emphasis on object-centric representations (such as object states in our case) can underperform for our task.

\begin{figure*}[t]
    \centering
    \includegraphics[width=\linewidth]{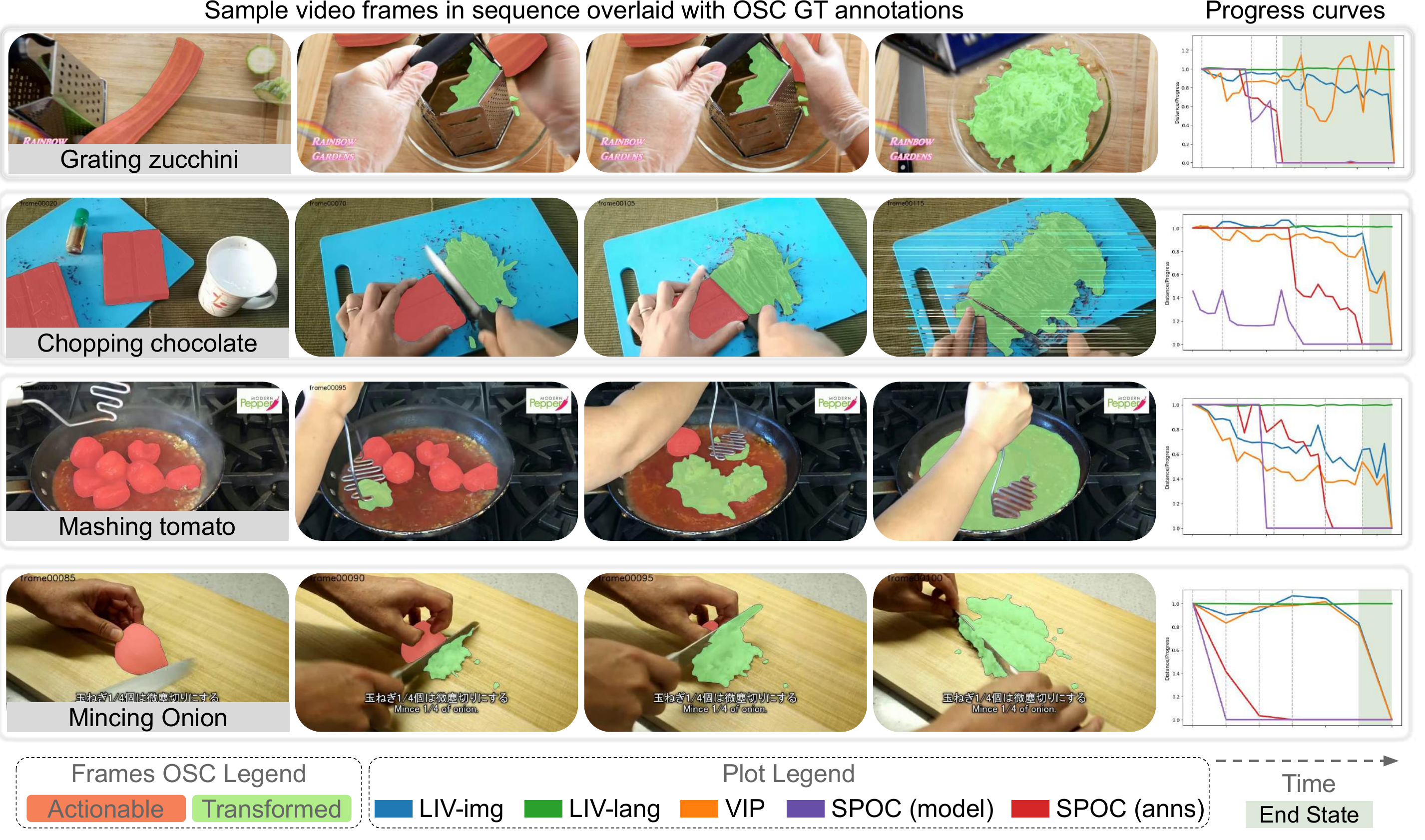}
    \caption{\textbf{Activity progress curves. } 
    Additional activity progress curves, like those in Fig.~\ref{fig:progress} in main. We show sample frames from a video sequence with progress curves generated by different methods, where vertical lines indicate the time-steps of sampled frames. 
    Ideal curves should decrease monotonically, and saturate upon reaching the end state. 
    In contrast to goal-based representation learning methods such as VIP~\cite{ma2022vip} and LIV~\cite{ma2023liv}, OSC-based curves accurately track task progress, making them valuable for downstream applications like progress monitoring and robot learning.
    }
    \label{fig:progress_supp}
    \vspace{1em}
\end{figure*}

In addition to the curves in Fig.~\ref{fig:progress} in main, we show additional activity progress curves in Fig.~\ref{fig:progress_supp}. We observe that OSC-based SPOC curves are generally more monotonic for the duration of the OSC, while the baselines tend to oscillate and remain stagnant for the most part. 

\custompar{End-state Sigma and L2}
This metric evaluates curve behavior once the activity is complete. Ideally, once the end state is reached and the task is complete, we would not like to observe any fluctuations in the progress curve. To measure this, we compute both the variance of the curve ($end_{\sigma}$) and its absolute L2 value ($end_{l2}$) in the end-state period. HowToChange has manual ground-truth annotations for classifying each frame in the clip into one of: {initial, transitioning, end-state}. We compute this metric in GT end-state annotated frames. Ideal values for both should be 0.

In Table~\ref{tab:progress} of the main paper, OSC-based curves have lower $end_{\sigma}$ and $end_{l2}$ values compared to the baselines. This is corroborated by the qualitative figures in Figs.~\ref{fig:progress} and ~\ref{fig:progress_supp}, where the OSC-based curves stagnate in the end-state. In contrast, the goal-based baselines, being sensitive to the goal image, stagnate during the actual activity progression, while rapidly decreasing in the end-state after the task is complete.
In certain scenarios where part of the object may be occluded (e.g. in Fig.~\ref{fig:progress_supp}: mincing onion, hand occludes whole onion), OSC-based curves can show early saturation. The baselines nevertheless fail to track progress.

\section{Limitations and Future Work}
\label{sec_supp:limitations}

\begin{figure*}[t]
    \centering
    \includegraphics[width=\linewidth]{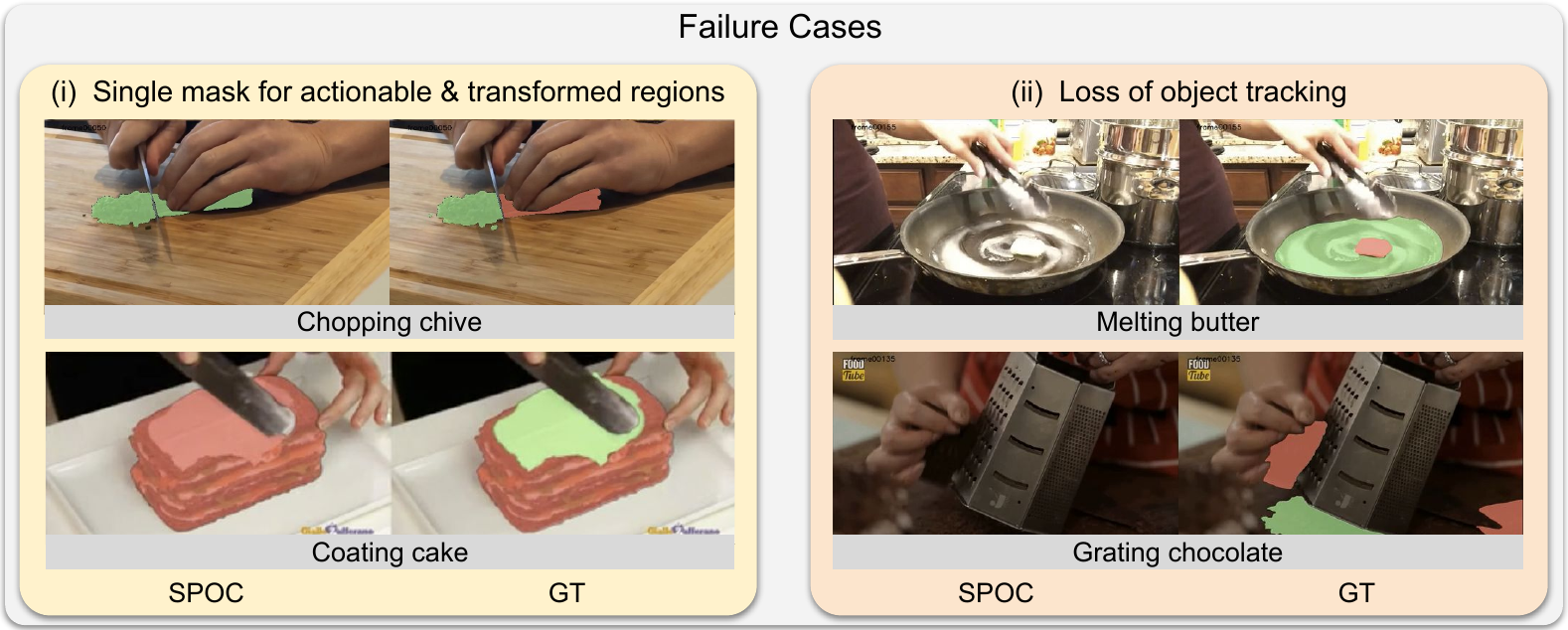}
    \caption{\textbf{Failure modes. } 
    We observe two main modes of failure in our model: \textbf{(i)} single mask proposal for both actionable and transformed regions (e.g. chopped and unchopped chive regions are within a single mask), and \textbf{(ii)} loss of object tracking (e.g. butter ceases to be tracked due to rapid motion). We discuss these cases in detail and discuss solutions in Sec.~\ref{sec_supp:limitations}.
    }
    \label{fig:failure}
\end{figure*}

In addition to depicting failure cases in Fig.~\ref{fig:main_results}d of the main paper, here we discuss them in detail. We observe two main modes of failure in our model: single mask proposal for both actionable and transformed regions, and loss of object tracking. Samples from each failure mode are shown in Fig.~\ref{fig:failure}. We explain each case below.

The first bottleneck in our method is the detector we adopt to identify the object of interest. Existing open-world detectors have been trained to output bounding boxes for the whole object region. As a result, we find that when the actionable and transformed regions are very close to each other (e.g. Fig.~\ref{fig:failure} (i): chopping chive), the detector can output a single bounding box including both regions. As a result, we obtain a single label for both actionable and transformed regions, limiting the intra-object capability of our model (Table~\ref{tab:htc_abl}-Transition). This failure mode is less pronounced when there is a meaningful separation between the two regions (e.g. grating carrot, where the whole carrot and grated pieces are separated).

A second bottleneck for our method is the mask tracker we rely on for tracking detected masks in successive time-steps. Existing trackers work well for objects that remain static through time. Our dataset is primarily comprised of dynamically changing objects that often evolve drastically in terms of size, shape, texture, color and so on. This dynamic object morphing proves to be a challenge for existing trackers. As a result, we notice that tracked masks can sometimes taper out if the object is undergoing fast motions or transitions (e.g. Fig.~\ref{fig:failure} (ii): melting butter). As a result, running the detector at regular intervals becomes important, to offset some of the false negatives. The effect of this failure mode is more pronounced for WTC-VOST subset, which comprises continuously captured in-the-wild egocentric videos. Sudden, jerky head motions common in egocentric videos often throw off the tracker, leading to reduced segmentation performance.

Third, our baseline trains a separate model per verb, limiting generalization: unseen actions would require new training data. A promising future direction is to develop a multi-task verb-conditioned model, enabling transfer to unseen actions and objects with minimal supervision.

Future advancements in object detection and mask tracking are likely to further enhance our model's performance and alleviate the issues caused by these two failure modes. In addition, an exciting avenue to explore would be to predict mask proposals for actionable and transformed regions directly from the image. Such a method would have the dual benefit of learning intra-object mask proposal end-to-end while also being sensitive to OSC dynamics over time.


\end{document}